\documentclass[10pt,twocolumn,letterpaper]{article}

\usepackage{wacv}
\usepackage{times}
\usepackage{epsfig}
\usepackage{graphicx}
\usepackage{amsmath}
\usepackage{amssymb}

\usepackage{lipsum}
\usepackage{xcolor}
\usepackage{booktabs}
\usepackage{tikz}
\usepackage{multirow}

\newcommand{\myparagraph}[1]{\textbf{#1}\quad}

\makeatletter
\newcommand\footnoteref[1]{\protected@xdef\@thefnmark{\ref{#1}}\@footnotemark}
\makeatother

\def\wacvPaperID{1155} 

\wacvfinalcopy 
\ifwacvfinal
\fi

\ifwacvfinal
\usepackage[breaklinks=true,bookmarks=false]{hyperref}
\else
\usepackage[pagebackref=true,breaklinks=true,colorlinks,bookmarks=false]{hyperref}
\fi

\newcommand{\teaser}{
    \centering
    \includegraphics[width=.29\textwidth]{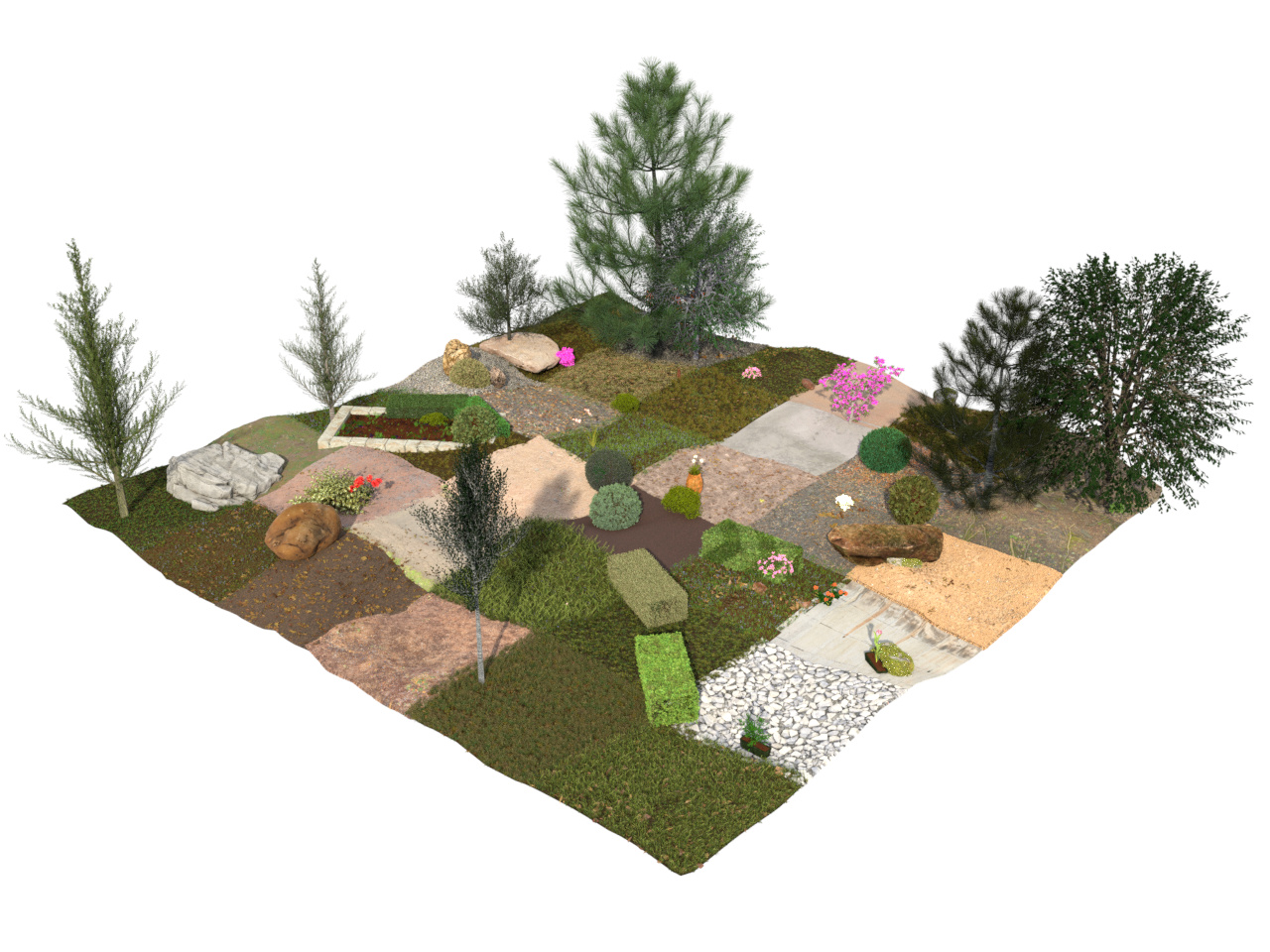}~
    \includegraphics[width=.29\textwidth]{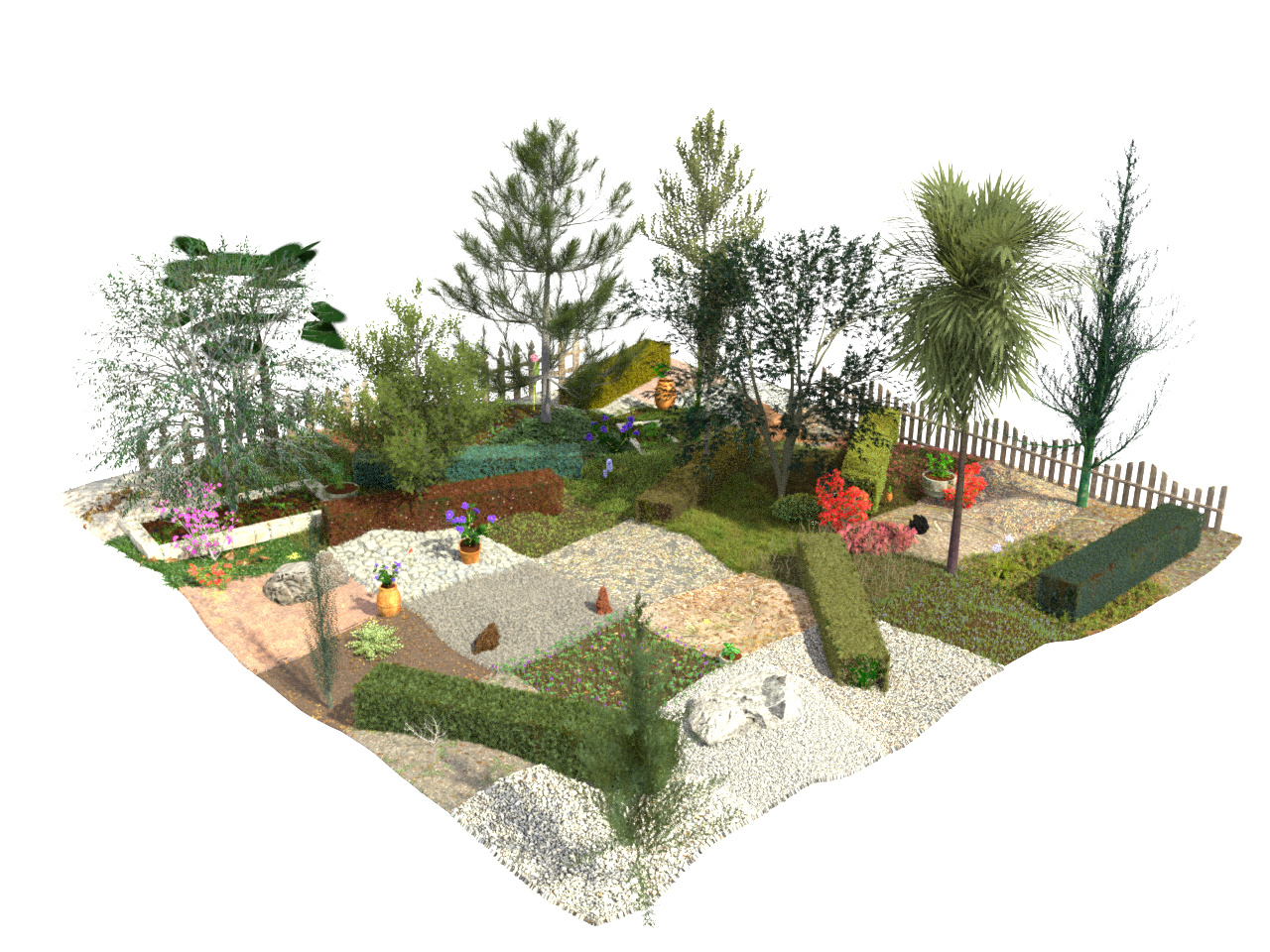}~
    \includegraphics[width=.29\textwidth]{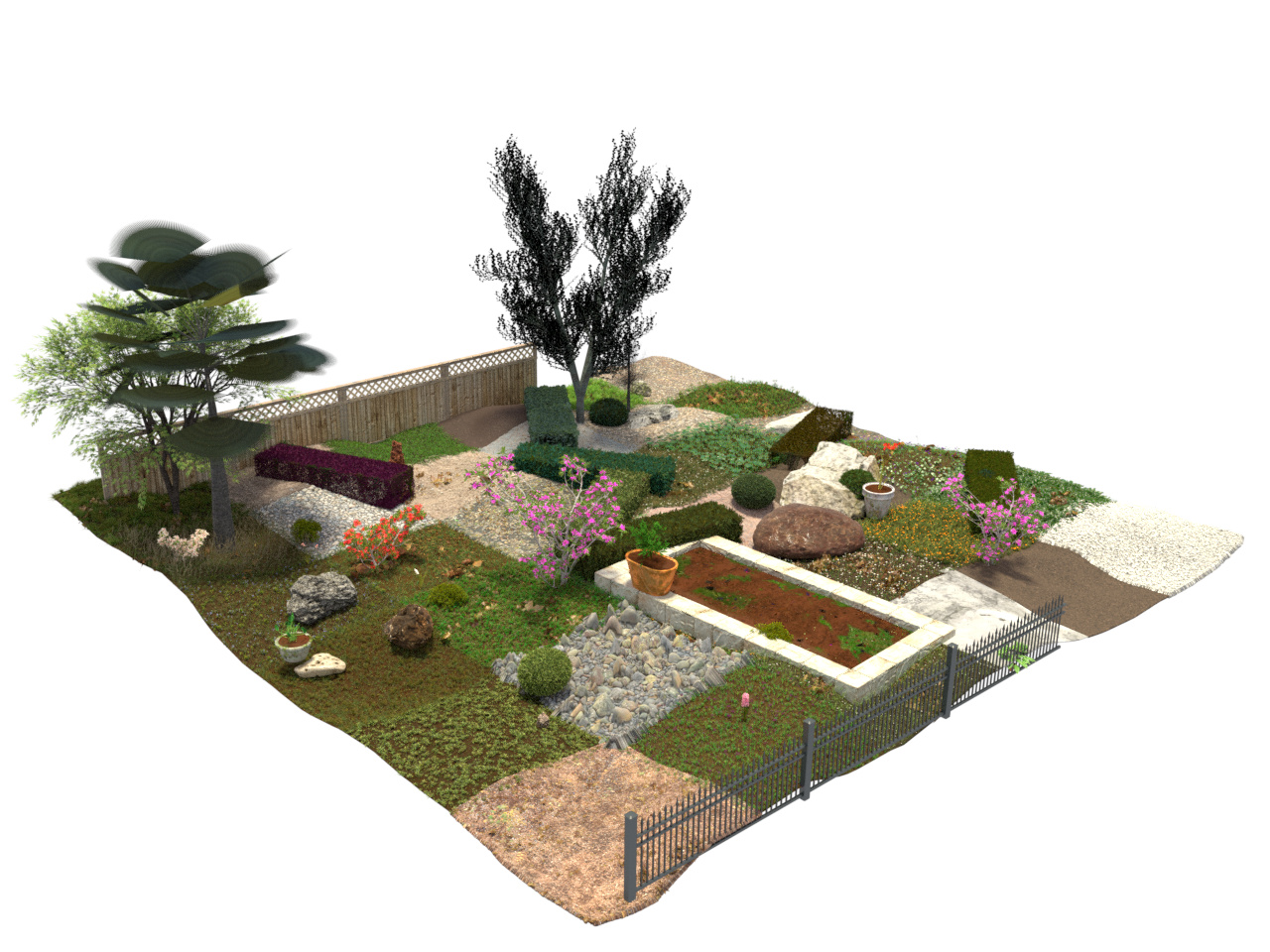} \\
    \includegraphics[width=.29\textwidth]{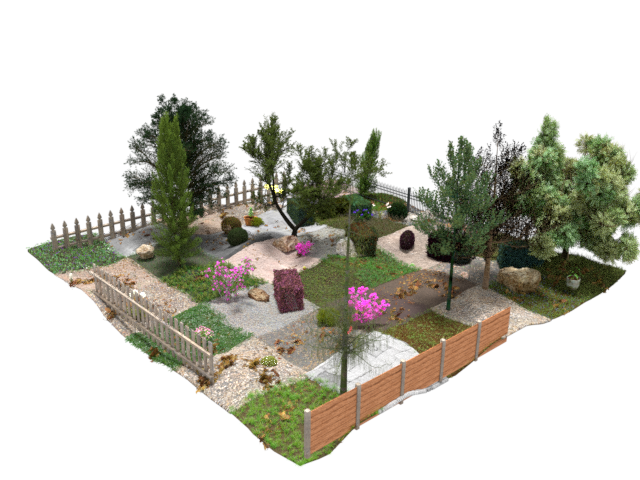}~
    \includegraphics[width=.29\textwidth]{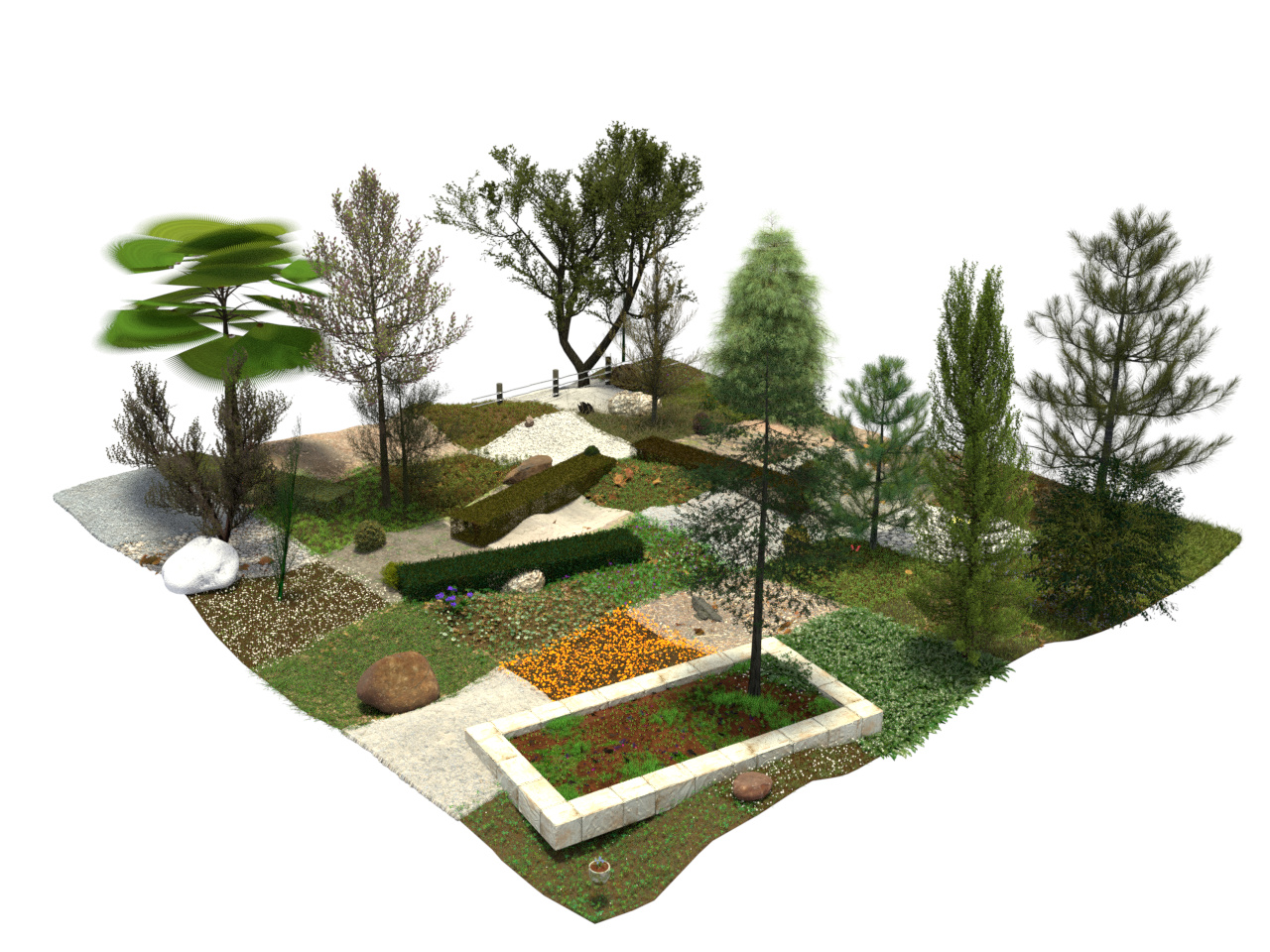}~
    \includegraphics[width=.29\textwidth]{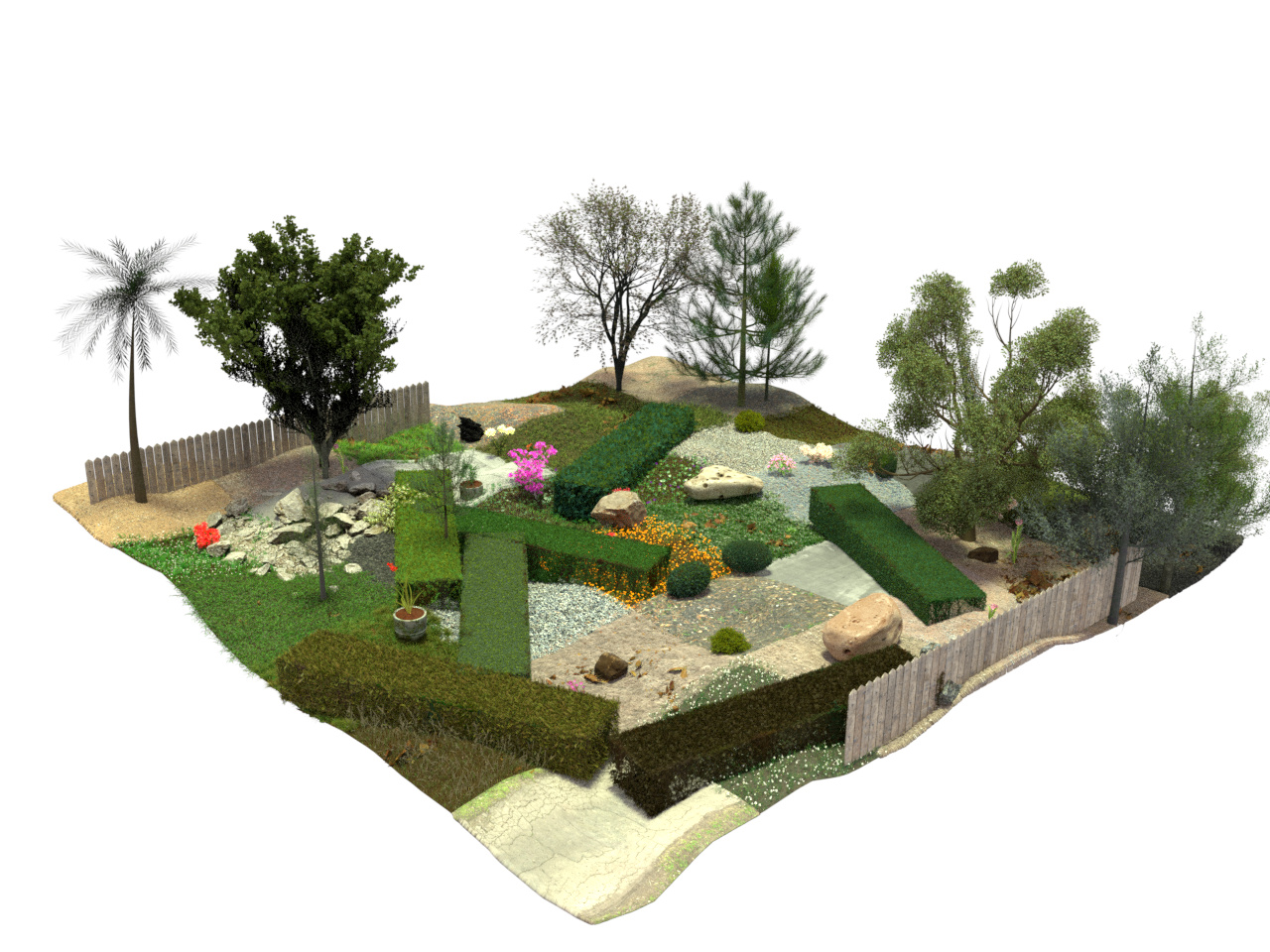}~
}

\makeatletter
\let\@oldmaketitle\@maketitle
\renewcommand{\@maketitle}{\@oldmaketitle
    \teaser\bigskip
    }
\makeatother

\begin{document}

\title{EDEN: Multimodal Synthetic Dataset of \underline{E}nclosed Gar\underline{DEN} Scenes}

\author{

Hoang-An Le$^{1}\quad$ Thomas Mensink$^{1,2}\quad$ Partha Das$^{1,3}\quad$ Sezer Karaoglu$^{1,3}\quad$ Theo Gevers$^{1,3}$
\\ $^{1}$Computer Vision lab, University of Amsterdam
\\ $^{2}$Google Research Amsterdam $\qquad ^{3}$3DUniversum, Amsterdam \\ 
{\tt\small \{h.a.le, p.das, s.karaoglu, th.gevers\}@uva.nl, mensink@google.com}

}

\maketitle

\graphicspath{{Images/}}

\begin{abstract}

Multimodal large-scale datasets for outdoor scenes are mostly designed for urban
driving problems. The scenes are highly structured and semantically different from
scenarios seen in nature-centered scenes such as gardens or parks. To promote
machine learning methods for nature-oriented applications, such as agriculture
and gardening, we propose the multimodal
synthetic dataset for Enclosed garDEN scenes (EDEN). The dataset features more than 300K
images captured from more than 100 garden models. Each image is annotated with various
low/high-level vision modalities, including semantic segmentation, depth, surface normals,
intrinsic colors, and optical flow. Experimental results on the state-of-the-art methods
for semantic segmentation and monocular depth prediction, two important tasks in
computer vision, show positive impact of pre-training deep networks on our dataset
for unstructured natural scenes. The dataset and related materials will be available at \url{https://lhoangan.github.io/eden}.

\end{abstract}
\section{Introduction}
\label{sec:Intro}

Synthetic data have been used to study a wide range of 
computer vision problems since the early days~\cite{Aggarwal1988,Barron1994,Horn1981}.
Compared to real-world imagery (RWI), computer-generated imagery (CGI) 
data provides allows for less expensive and more accurate annotation.
Since the emergence of deep learning, synthetic datasets using CGI has become essential due to the data-hungry nature of deep learning methods and the difficulty of annotating real-world images. Most of the large-scale RWI datasets (with more than 20K annotated data points) are focusing on higher-level computer vision tasks such as (2D/3D) detection, recognition, and segmentation~\cite{matterport3d2017,scannet2017,imagenet,mscoco,mapillary2017,ade20k}. In contrast, datasets for low-level image processing such as optical flow, visual odometry (KITTI~\cite{kitti2012,kitti2015}) and intrinsic image decomposition (IIW~\cite{iiw2014}, MIT~\cite{mit2009},  SAW~\cite{saw2017}) are limited in the number of samples (around 5K annotated images).

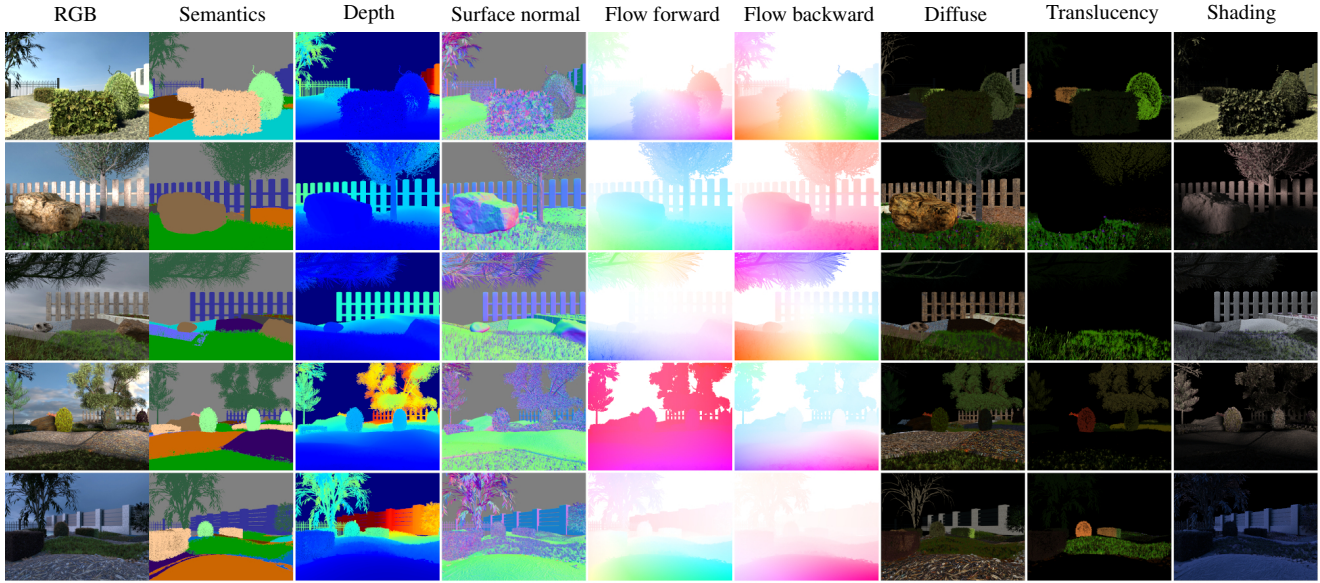
\begin{figure*}[t]
    \centering
    \begin{tikzpicture}
        \node {
            \includegraphics[width=\textwidth]{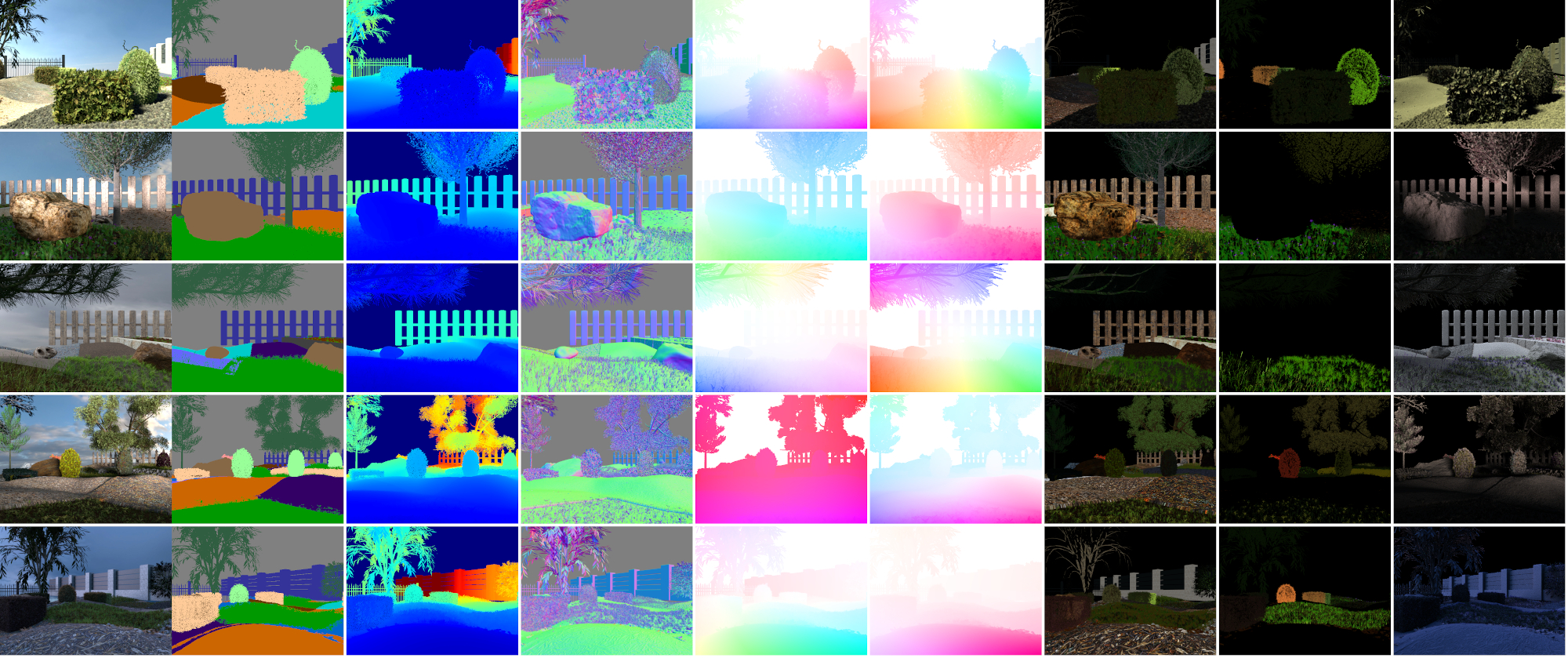}
        };
        \node at (-7.80,3.9) {\footnotesize RGB};
        \node at (-5.85,3.9) {\footnotesize Semantics};
        \node at (-3.90,3.9) {\footnotesize Depth};
        \node at (-1.95,3.9) {\footnotesize Surface normal};
        \node at ( 0.00,3.9) {\footnotesize Flow forward};
        \node at ( 1.95,3.9) {\footnotesize Flow backward};
        \node at ( 3.90,3.9) {\footnotesize Diffuse};
        \node at ( 5.85,3.9) {\footnotesize Translucency};
        \node at ( 7.70,3.9) {\footnotesize Shading};
    \end{tikzpicture}
    \caption{An overview of multiple data types provided in the dataset. The dataset includes data for both low- and high-level tasks such as (stereo) RGB, camera odometry, instant and semantic segmentation, depth, surface normal, forward and backward optical flow, intrinsic images (albedo, shading for diffuse materials, translucency)}
    \label{fig:rendered_data}
\end{figure*}

CGI-based synthetic
datasets~\cite{sintel2012,vkitti2016,gtav2018,flyingthings3d2016,viper2017,synthia2016} provide more and diverse annotation types.
The continuous progress of computer graphics and video-game industry results in improved photo-realism in render engines. The use of physics-based renderers facilitates the simulation of scenes under different lighting conditions (\eg morning, sunset, nighttime).
Information
obtained by video-game pixel shaders \cite{gtav2018,viper2017,Richter2016} is of high-quality and can be used to train low-level computer vision tasks such as
optical flow, visual odometry and intrinsic image decomposition.

Most of the existing datasets focus on car driving scenarios and are mostly composed of simulations of urban/suburban scenes. City scenes are structured containing objects that are geometrically distinctive with clear boundaries. However, natural or agriculture scenes are often unstructured.
The gaps between them are large and required distinctive attentions. For example, 
there are only trails and no drive ways nor lane marks for travelling; bushes and
plants are deformable and often entangled; obstacles such as small boulders may 
cause more trouble than tall grass.

To facilitate the development of computer vision and (deep)
machine learning for farming and gardening applications, which involve mainly
unstructured scenes, in this paper, we propose the synthetic dataset of Enclosed
garDEN scenes (EDEN), the first large-scale  multimodal dataset with $>$300K
images, containing a wide range of botanical objects (\eg trees, shrubs, flowers),
natural elements (\eg terrains, rocks), and garden objects (hedges, topiaries). The
dataset is created within the TrimBot2020
project\footnote{ \url{http://trimbot2020.webhosting.rug.nl/}}
for gardening robots, and have pre-released versions used in the
3DRMS challenge~\cite{3drms2018} and in several
work~\cite{Baslamisli2019arxiv,Baslamisli18eccv,HALe2018}.

In contrast to man-made (structured) objects in urban scenarios (such as buildings,
cars, poles, \etc), the modelling of natural (unstructured) objects is more
challenging. Natural objects appear with their own patterns and shapes, making a
simplified or overly complex object easily recognized as unrealistic. Rendering
techniques using rotating billboards of real photos may provide realistic
appearances, but lack close-up geometrical features. Although synthetic datasets
and video-games may offer natural objects and scenes, they often come with generic
labels (e.g. tree, grass, and simple vegetation), since their focus is on the
gaming dynamics.
Therefore, objects in our dataset are developed using high-fidelity parametric
models or CADs created by artists to obtain natural looking scenes. The object
categories are selected for the purpose of gardening and agricultural scenarios
to include a large variety of plant species and terrain types. The dataset contains
relatively different lighting conditions to simulate the intricate aspects of outdoor
environments. The different data modalities are useful for both low- and high-level
computer vision tasks.

In addition to the new dataset itself, we provide analyses and benchmarks of the
dataset on state-of-the-art methods of two important tasks in computer vision,
namely semantic segmentation and depth prediction.

\begin{figure*}[t]
    \centering
    \includegraphics[width=.9\textwidth]{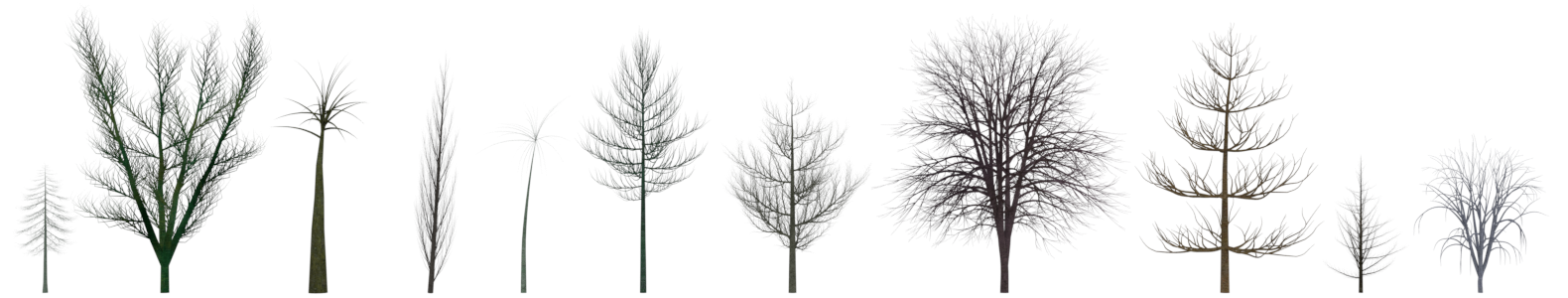}
    \includegraphics[width=.9\textwidth]{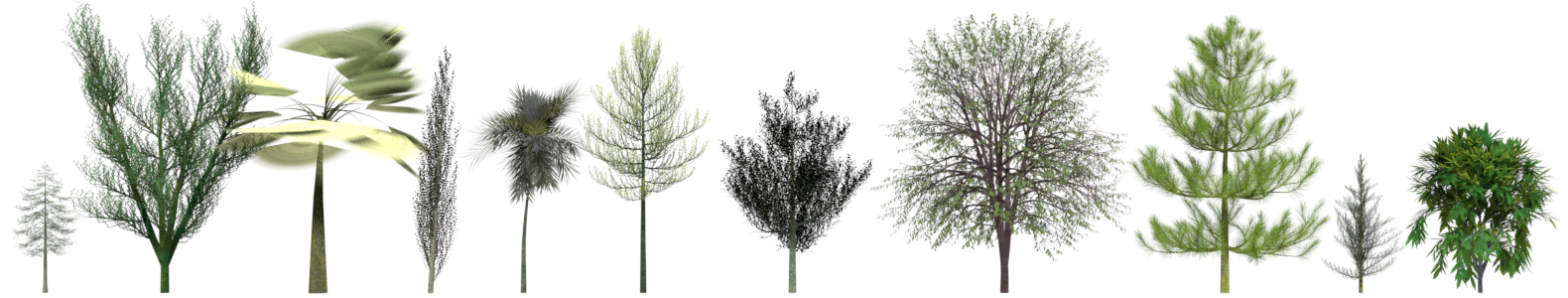}
    \caption{Sample tree models (top: tree stems, bottom: with leaves) for various tree species}
    \label{fig:tree_sample}
\end{figure*}

\section{Related Work}
\label{sec:related_work}

\subsection{Real-imagery datasets}
\label{subsec:real_datasets}

To accommodate the emergence of deep learning and its data-demanding nature, many efforts have been spent on creating large-scale generic datasets, starting with the well-known
ImageNet~\cite{imagenet}, COCO~\cite{mscoco}, and 
Places~\cite{places2014}. These are real-world imagery (RWI) datasets with more than 300K annotated images at object and scene-level. Also in the domain of semantic segmentation, there are a number of datasets available such as Pascal-Context~\cite{pascalContext} (10,103 images, 540 categories) and ADE20K~\cite{ade20k} (20,210 images, 150 categories).

Annotation is expensive. Lower-level task annotation is even more expensive.
In contrast to the availability of large datasets for higher-level computer vision tasks, there are only a few RWI datasets for low-level tasks such as optical flow, visual odometry, and intrinsic image decomposition due unintuitive data annotation. Middlebury~\cite{middlebury2011} and
KITTI~\cite{kitti2012,kitti2015} are the only datasets providing optical flow for real-world images, yet too small to train a deep network effectively. For intrinsic
image decomposition, the MIT~\cite{mit2009} dataset provides albedo and shading ground truths for only 20 objects in controlled lighting conditions, while IIW~\cite{iiw2014} and SAW~\cite{saw2017} provide for up to 7K in-the-wild and indoor images. Indoor-scene datasets~\cite{nyu2012,stanford23d2017arxiv,matterport3d2017,scannet2017} provide a larger number of images (up to 2.5M) and with more modalities (such as depth) than generic datasets. However, their goal is to provide data for 3D (higher-level) indoor computer vision tasks.

Outdoor scenes are subject to changing imaging conditions, such as lighting conditions, viewpoint, occlusion and object appearance, resulting in annotation difficulties. A number of methods are proposed focusing on scene understanding for autonomous driving~\cite{Leibe2007,camvid2008,kitti2012,kitti2015,cityscapes2016,mapillary2017}. However, these datasets are limited in number of images and/or the number modalities. Mapillary~\cite{mapillary2017,mapillary2020} is the most diverse dataset with varying illumination conditions, city views, weather and seasonal changes. Their focus is on semantic segmentation and place recognition. Large-scale multimodal datasets are restricted to synthetic data.

\subsection{Synthetic datasets}
\label{subsec:synthetic_datasets}
Computer vision research uses synthetic datasets since the early days
to study low-level tasks,~\eg optical flow~\cite{Horn1981,Aggarwal1988,Barron1994}.
Synthetic data provide cheaper and more accurate annotations. It can facilitate noise-free and controlled environments for otherwise costly
problems~\cite{Taylor2007,Mueller2016} or for intrinsic understanding~\cite{Kaneva2011} and proof of concept~\cite{Kicanaoglu2018,Olszewski2019}.

Obviously, the quality of synthetic data and annotation depends on the realism of modelling and rendering algorithms. 
The development of computer graphic techniques has led to physics-based render engines and the improvement of photo-realistic computer-generated imagery (CGI). SYNTHIA~\cite{synthia2016} and
Virtual KITTI~\cite{vkitti2016} simulate various daylight conditions
(morning, sunset), weather (rain, snow), and seasonal variations (spring, summer, fall, winter) for autonomous (urban) driving datasets. Datasets obtained from
video-games~\cite{Richter2016,viper2017,gtav2018}
and movies~\cite{sintel2012,flyingthings3d2016} show adequate
photo-realism. These datasets provide not only dense annotations for  high and low-level tasks, but also images are taken from multiple
viewpoints and under different illumination/weather/seasonal settings. They have proven useful for training robust deep models under different environmental conditions~\cite{viper2017,gtav2018}.

\begin{figure*}[t]
    \centering
    \includegraphics[width=.9\textwidth]{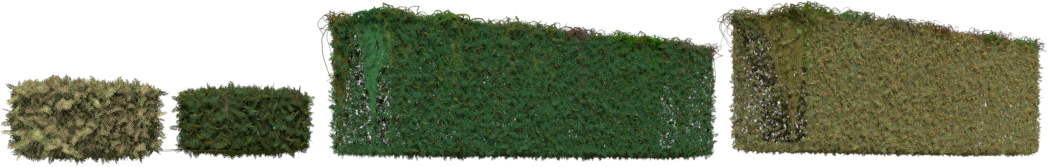}
    \includegraphics[width=.9\textwidth]{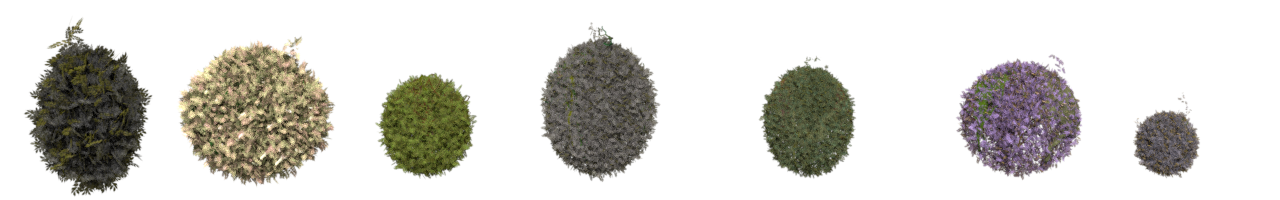}
    \caption{Sample models for hedges (top) and topiaries (bottom). The bushes can be generated with various sizes, leaf colors, and internal stem structures. }
    \label{fig:ivy_sample}
\end{figure*}

Datasets for outdoor scenes, real or synthetic, focus mostly on either generic or urban driving scenarios. They mainly consist of scenes containing man-made (rigid) objects, such as lane-marked streets, buildings, vehicles,~\etc.
Only a few datasets contain (non-rigid) nature environments (e.g. forests or gardens~\cite{3drms2018,freiburgforest2016}).

CGI-based datasets rely on the details of object models, and
computer-aided designed (CAD) model repositories, such as
ShapeNet~\cite{shapenet2015}, play an important role in urban driving
datasets~\cite{vkitti2016,synthia2016}. However, the models usually include rigid objects with low fidelity. Others focus on capturing the uniqueness of living entities, such as
humans~\cite{smpl2015,Han2018}, and trees~\cite{Weber1995,Hewitt2017,Barth2018}
to generate highly detailed models with realistic variations. 
Synthetic garden datasets have been used in
\cite{Baslamisli18eccv,HALe2018,3drms2018},  albeit these datasets are relatively small and have just one or two modalities and are not all publicly available.
In this paper, we use different parametric models, \eg~\cite{Weber1995}, to generate different botanical objects in an garden. We create multiple gardens with different illumination conditions, and extract multi-modal data (including RGB, semantic segmentation, depth, surface normals \etc) from each frame, yielding over 300K garden frames, which we will make publicly available.


\begin{figure*}[t]
    \centering
    \includegraphics[width=\textwidth]{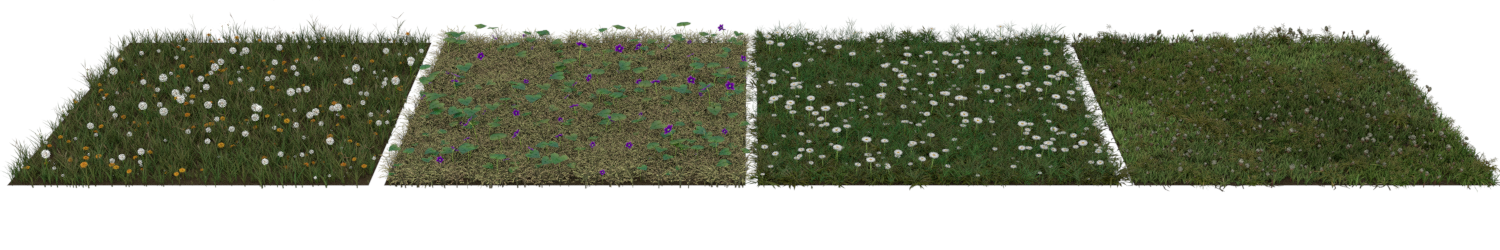}
    \includegraphics[width=\textwidth]{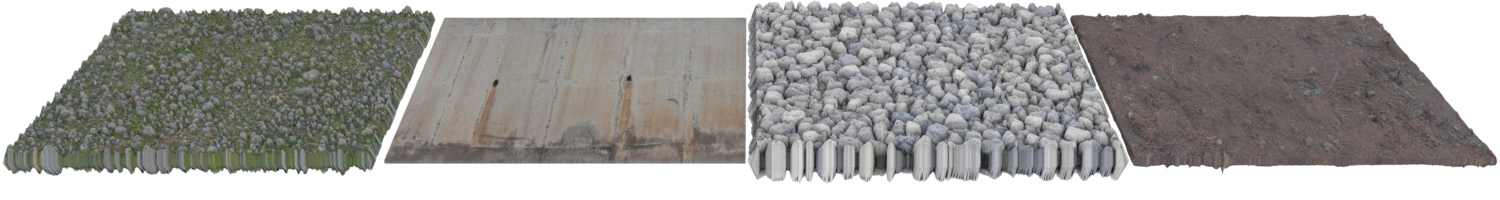}
    \caption{Sample tiles of different terrain types: grass with weed (\textit{top}), gravel, pavement, pebble stones, dirt (\textit{bottom}). The grass and weed
    species are chosen and combined randomly.
    }
    \label{fig:terrain}
\end{figure*}

\begin{figure*}[t]
    \centering
    \begin{tikzpicture}
        \node {
        \includegraphics[width=.19\textwidth]{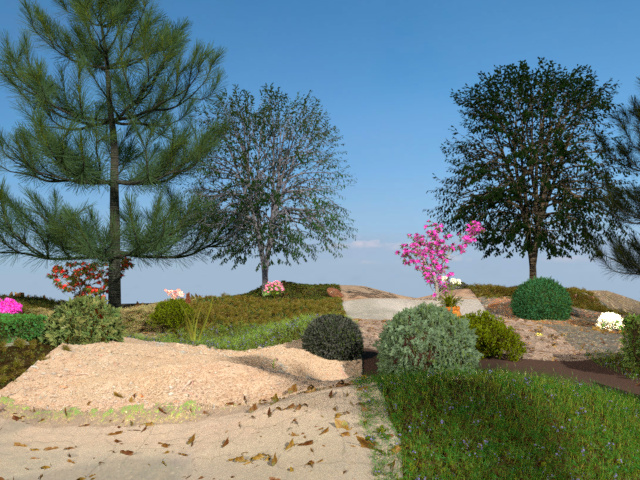}
        \includegraphics[width=.19\textwidth]{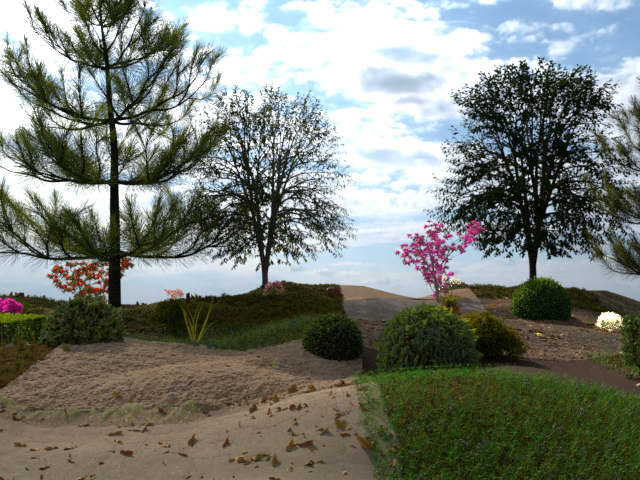}
        \includegraphics[width=.19\textwidth]{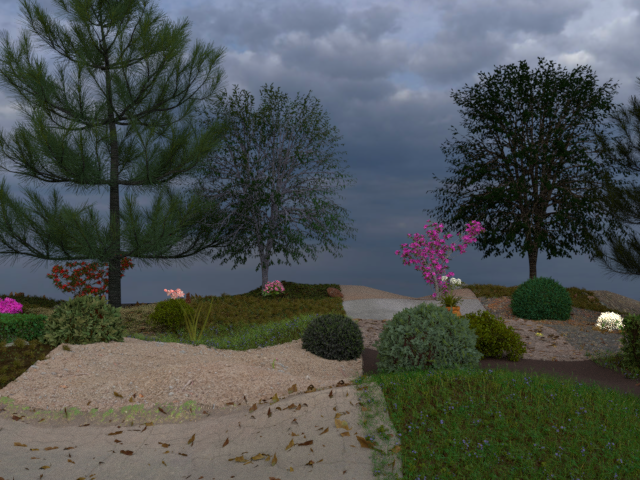}
        \includegraphics[width=.19\textwidth]{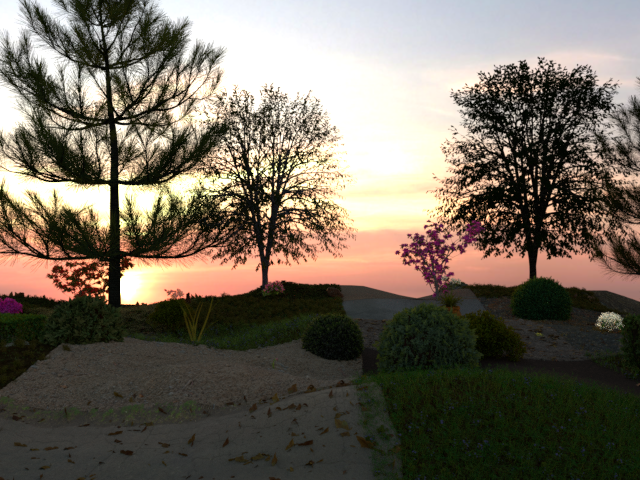}
        \includegraphics[width=.19\textwidth]{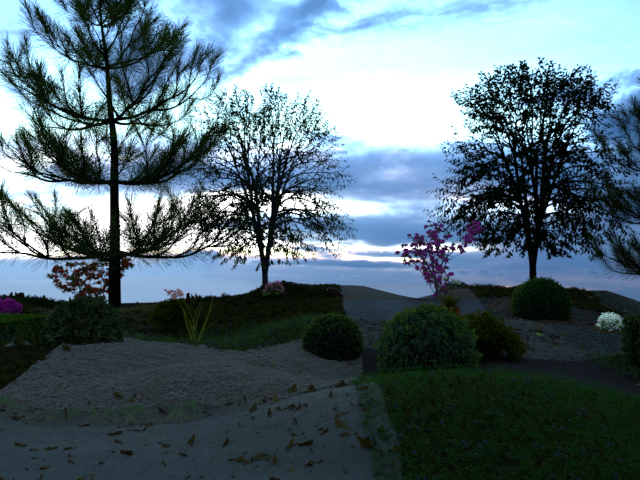}
        };
        \node at (-6.80,1.4) {\footnotesize clear};
        \node at (-3.40,1.4) {\footnotesize cloudy};
        \node at (    0,1.4) {\footnotesize overcast};
        \node at ( 3.40,1.4) {\footnotesize sunset};
        \node at ( 6.80,1.4) {\footnotesize twilight};
    \end{tikzpicture}
    \caption{Illustration for scene appearance changed according to different illumination conditions.}
    \label{fig:environment}
\end{figure*}

\section{Dataset Generation}
\label{sec:method}

We create synthetic gardens using the free and open-source software
of Blender\footnote{\url{blender.com}, GPL GNU General Public License version 2.0},
and render using the physics-based Cycles render engine. Each garden consists of a ground with different terrains and random objects (generated with random parameters or randomly chosen from a pre-designed models). The modelling details of each component object and the rendering settings are presented in the following sections.

\subsection{Modelling}
\label{subsec:param_model}

To expand the diversity of objects and scenes, we propose to
combine parametric and pre-built models in the generation process.

\myparagraph{Trees} We use the tree parametric model described in~\cite{Weber1995},
implemented by the Blender Sapling 
Add-on\footnote{\label{ft:link}See Sec.~\ref{sec:links} for the reference link}.
A tree is constructed recursively from common predefined tree shapes
(conical, (hemi-)spherical, (tapered) cylindrical,~\etc) with the first level
being the trunk. The parameters define the branch features such as length,
number of splits, curvatures, pointing angles,~\etc, each with a variation range for
random sampling. Leaves are also defined in a similar manner as stems, besides
a fractional value determining their orientation to simulate phototropism.
The model can generate different tree species such as quaking aspens, maples,
weeping pillows, and palm trees. We use the parameter presets provided in the sampling add-on
and Arbaro\footnoteref{ft:link} (Figure~\ref{fig:tree_sample}). Totally there are 19
common tree species.

\myparagraph{Bushes} Hedges and topiaries are generated by growing an ivy adhering to
a rectangular or spherical skeleton object using the Ivy
Generator\footnoteref{ft:link}, implemented
by the Blender IvyGen add-on\footnoteref{ft:link}.
An ivy is recursively generated from a single root point by forming curved objects under
different forces including a random influence to allow overgrowing, 
an adhesion force to keep it attached to the trellis, a gravity pulling down, and an up-vector simulating phototropism. The add-on is known for creating realistic-looking ivy objects
(Figure~\ref{fig:ivy_sample}). We use more than 20 leaf types with different color augmentation for
both topiaries and hedges.

\myparagraph{Landscapes and terrain}
The landscape is created from a subdivided plane using a displacement modifier with the Blender cloud gradient noise which is a representation of Perlin noise~\cite{Perlin1985}. The modifier
displaces each sub-vertex on the plane according to the texture intensity, creating the
undulating ground effect. The base ground is fixed at 10x10 square meters, on which are paved
the terrain patches of 1x1 square meter. Each patch is randomly assigned to one of the terrain
types, including grass, pebble stones, gravels, dirt and pavement.

The grass is constructed using Blender particle modifier which replicates a small number of
elemental objects, known as particles, over a surface. We use the grass particles provided by
the Grass Essentials\footnoteref{ft:link}, and
the Grass package\footnoteref{ft:link}, containing expert-designed realistic-looking grass particles.
There are more than 30 species of
grass,~\eg St. Augustine grass, bahiagrass, centipedegrass,~\etc
and weed,~\eg dandelions, speedwell, prickly lettuce,~\etc. Each species has up to 49 model
variations. The appearance of the grass patch is controlled via numerical parameters,
such as freshness, brownness, wetness, trimmed levels, lawn stripe shape for mowed field,~\etc.
Illustrations for different grass and weed species are shown in Figure~\ref{fig:terrain} (top).

The other terrains are designed using textures from the Poliigon
collection\footnoteref{ft:link} of
high quality photo-scanned textures. Illustrations are shown in Figure~\ref{fig:terrain} (bottom).
Each texture contains a reflectance, surface normal, glossy, and reflection map with
expert-designed shaders for photo-realism. The resulted landscapes can be seen
on the first page.

\myparagraph{Environment}
Lighting in our dataset is created by 2 sources, a sun lamp
and a sky texture. A sun lamp is a direct parallel light source, simulating an
infinitely far light source. The source parameters include direction, intensity,
size (shadow sharpness), and color. A sky texture provides the environmental
background of rendered images and a source of ambient lights. We use the 
Pro-Lighting: Skies package\footnoteref{ft:link} composing of 95 realistic equirectangular HDR sky images of
various illuminations. The images are manually chosen and divided into 5 scenarios,
namely clear (sky), cloudy, overcast, sunset, and twilight. We also use 76 HDR
scenery images\footnoteref{ft:link} to create more various and
complex backgrounds, some with night lighting, coined scenery. An example of
lighting effects is shown in Figure~\ref{fig:environment}.

\myparagraph{Pre-built models}
To enhance the model variations in the dataset, we also include models prebuilt 
from different artists, including 
rocks\footnoteref{ft:link},
flowers\footnoteref{ft:link},
garden assets such as fences, flower 
pots\footnoteref{ft:link},~\etc.

\myparagraph{Garden construction}
For each garden, 2 to 4 types are sampled of each
grass species, as well as for tree, terrains, bushes, rocks, flowers, and garden assets.
The number of tree, bush, and obstacle instances are uniformly sampled from the closed intervals
$[5,17], [10,24]$, and $[3, 17]$, respectively; each instance is randomly assigned with one of the
corresponding species.
The random seeds in parametric models allow plants of the same species to contain
internal variations. The objects are distributed
at random places around the garden, avoiding overlapping each others, while the fences,
if any, are placed at the 4 edges.

\subsection{Rendering}
\label{subsec:rendering}

\begin{figure}
    \centering
    \def\svgwidth{.55\linewidth}
    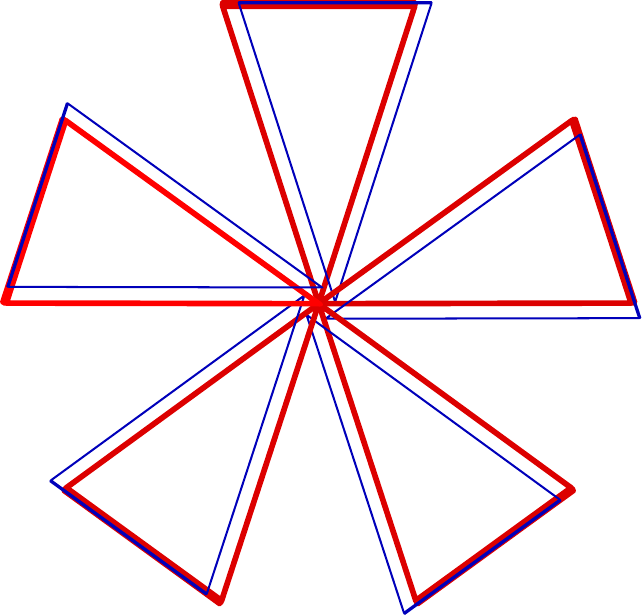
    \caption{The camera system: a ring of 5 pairs of stereo cameras at $72^\circ$ angular separation}
    \label{fig:cam_sys}
\end{figure}

\myparagraph{Camera setup}
We follow the real-world camera setup in the 3DRMS challenge to create a ring of 5 pairs of virtual
stereo cameras with angular separation of $72^\circ$ (Figure~\ref{fig:cam_sys}),
baseline of 0.03 meters. Each camera has a virtual
focal length of 32mm on a 32mm wide simulated sensor. The rendered resolution is
set to VGA-standard of 480x640 pixels. The camera intrinsic matrix is as follows:

\begin{equation}
    \bf K = \left[ {\begin{array}{*{20}{c}}
		   {640} & 0 & {320}  \\
		   0 & {640} & {240}  \\
		   0 & 0 & 1  \\
		\end{array}} \right].
\end{equation}

\begin{figure}[t]
    \centering
    \includegraphics[width=.23\textwidth]{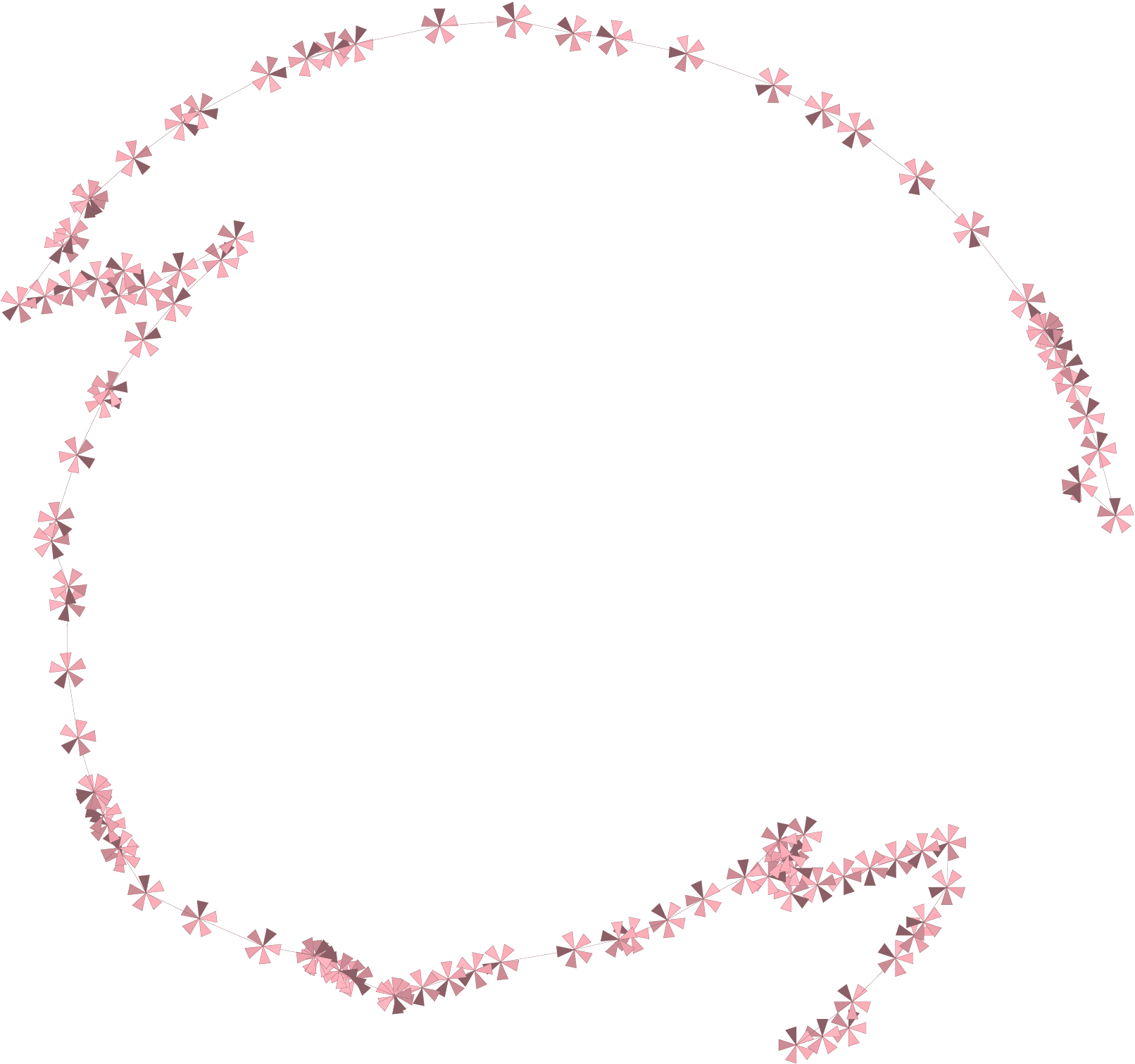}
    \includegraphics[width=.23\textwidth]{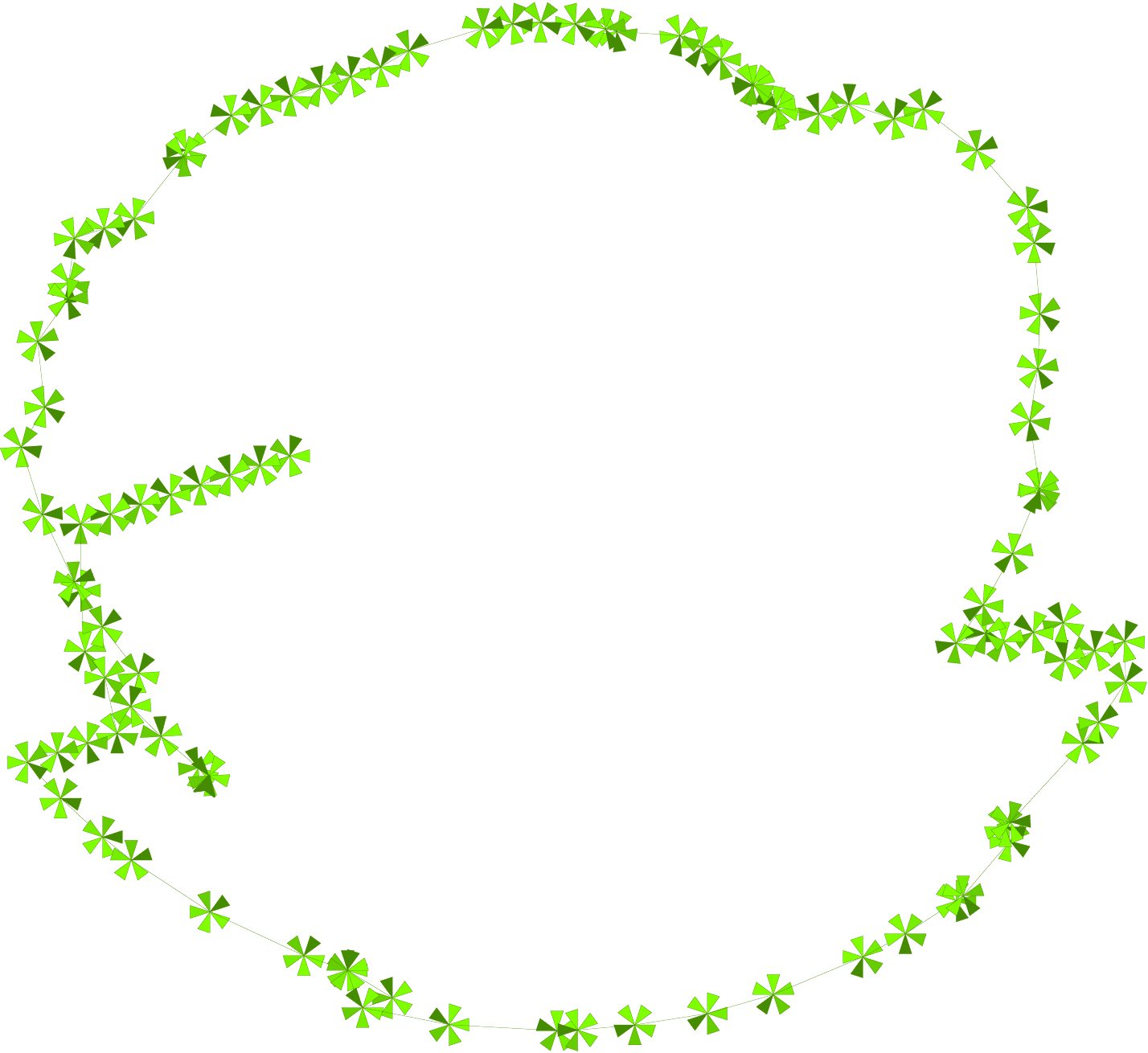}
    \caption{Examples of the generated trajectories used in the rendering process.
    The 5 pairs of cameras, illustrated by different color shades, are randomly
    moved, turned, and self-rotated while avoiding obstacles in a garden.
    }
    \label{fig:campath_single}
\end{figure}

We generate a random trajectory for the camera ring for each illumination variation
of each garden model. The speed is set to about $0.5\text{m/s}$, frame rate of $10fps$,
simulating a trimming robot in a garden. To improve the variability, the camera ring
is set to randomly turn after a random number of steps and avoid running through the
objects. The turning angles are also randomized to include both gradual and
abrupt angles. The trajectory lengths are set to be at least 100 steps. The examples are shown in Figure~\ref{fig:campath_single}.

\myparagraph{Render engine}
Blender Cycles is a probabilistic ray-tracing render engine that 
derives the color at each pixel by tracing the paths of light from the camera back
to the light sources. The appearances of the objects are determined by the
objects' material properties defined by the bidirectional scattering distribution
function (BSDF) shaders, such as diffuse BSDF, glossy BSDF, translucent BSDF,~\etc.

Scene aspects such as geometry, motion and the object material properties are
rendered into individual images before being combined into a final image. The
formation of a final image $I({\bf x})$ at position $\bf x$ is as
follows\footnote{\cf Blender 2.83 \href{https://docs.blender.org/manual/en/latest/render/layers/passes.html\#cycles}{Manual},
last access July 2020}:

\begin{align}
    f_g({\bf x }) &= g_\text{color}({\bf x}) (g_\text{direct}({\bf x}) + g_\text{indirect}({\bf x})), \\
    I({\bf x }) &= f_D({\bf x}) + f_G({\bf x}) + f_T({\bf x}) + B({\bf x}) + E({\bf x}),
\end{align}

\noindent where $D, G, T, B, E$ are respectively the diffuse, glossy,
transmission, background, and emission passes. $D_\text{color}$ is the object colors
returned by the diffuse BSDF, also known as albedo; $D_\text{direct}$ is the lighting
coming directly from light sources, the background, or ambient occlusion
returned by the diffuse BSDF, while $D_\text{indirect}$ after more than
one reflection or transmission off a surface. Similar are $G$ and $T$ with
glossy and transmission BSDFs. Emission and background are pixels from directly
visible objects and environmental textures. The intermediate image contains at each pixel 
the corresponding data or zeros otherwise.

All the computations are carried out in the linear RGB space. Blender converts
the composite image to sRGB space using the following gamma-correction formula
and clipped to $[0,1]$ before saving to disk:

\begin{align}
\gamma (u)={\begin{cases}12.92u \quad &u\leq 0.0031308\\
1.055u^{1/2.4}-0.055 \quad &{\text{otherwise}}\end{cases}}
\end{align}

In our dataset, besides the $RGB$ stereo pairs and cameras' poses, we provide the
images from intermediate stages, namely albedo, shading, glossy, translucency,~\etc
for the left camera. As the modelling and rendering are physics-based,
the intermediate images represent different real-life modalities, such as geometry,
motion, intrinsic colors,~\etc. Examples
are shown in Figure~\ref{fig:rendered_data}.


\begin{table}[t]
    \centering
    \begin{tabular}{@{}cccc@{}}
        \toprule
        \multirow{2}{*}{Split} & \multirow{2}{*}{train (127)} & \multicolumn{2}{c}{test (20) } \\
        \cmidrule(r){3-4} 
        & & full & 20K \\
        \midrule
        clear & 74,913  & 10,035 & 3,333\\
        cloudy & 73,785 & 10,030 & 3,378\\
        overcast & 73,260 & 10,015 & 3,349\\
        sunset & 73,715 & 10,040 & 3,250\\
        twilight & 73,990 & 10,045&3,369\\
        \midrule
        total &  369,663 &  50,165 & 20,000\\
        \bottomrule
    \end{tabular}
    \caption{Number of images per scene and split; the number of models are in parentheses}
    \label{tab:stats}
\end{table}

\section{Experiments}
\label{sec:experiments}

In this section, the goal is to quantitatively analyze the newly created dataset to assess its realism and usability.
The evaluation is performed
via two proxy tasks: semantic segmentation and monocular depth estimation.

We split the dataset into training (127 models, 369,663 monocular images)
and test set (20 models, 60,195 images). To speed up the evaluation process,
we uniformly sample 20K images from the full test set. The statistics are shown
in Table~\ref{tab:stats}. The sample list will also be released together with the dataset.

\subsection{Semantic segmentation}

\begin{figure}[t]
    \centering
    \def\svgwidth{\columnwidth}
\begingroup%
  \makeatletter%
  \providecommand\color[2][]{%
    \errmessage{(Inkscape) Color is used for the text in Inkscape, but the package 'color.sty' is not loaded}%
    \renewcommand\color[2][]{}%
  }%
  \providecommand\transparent[1]{%
    \errmessage{(Inkscape) Transparency is used (non-zero) for the text in Inkscape, but the package 'transparent.sty' is not loaded}%
    \renewcommand\transparent[1]{}%
  }%
  \providecommand\rotatebox[2]{#2}%
  \newcommand*\fsize{\dimexpr\f@size pt\relax}%
  \newcommand*\lineheight[1]{\fontsize{\fsize}{#1\fsize}\selectfont}%
  \ifx\svgwidth\undefined%
    \setlength{\unitlength}{438.61858842bp}%
    \ifx\svgscale\undefined%
      \relax%
    \else%
      \setlength{\unitlength}{\unitlength * \real{\svgscale}}%
    \fi%
  \else%
    \setlength{\unitlength}{\svgwidth}%
  \fi%
  \global\let\svgwidth\undefined%
  \global\let\svgscale\undefined%
  \makeatother%
  \begin{picture}(1,0.36193175)%
    \lineheight{1}%
    \setlength\tabcolsep{0pt}%
    \put(0,0){\includegraphics[width=\unitlength,page=1]{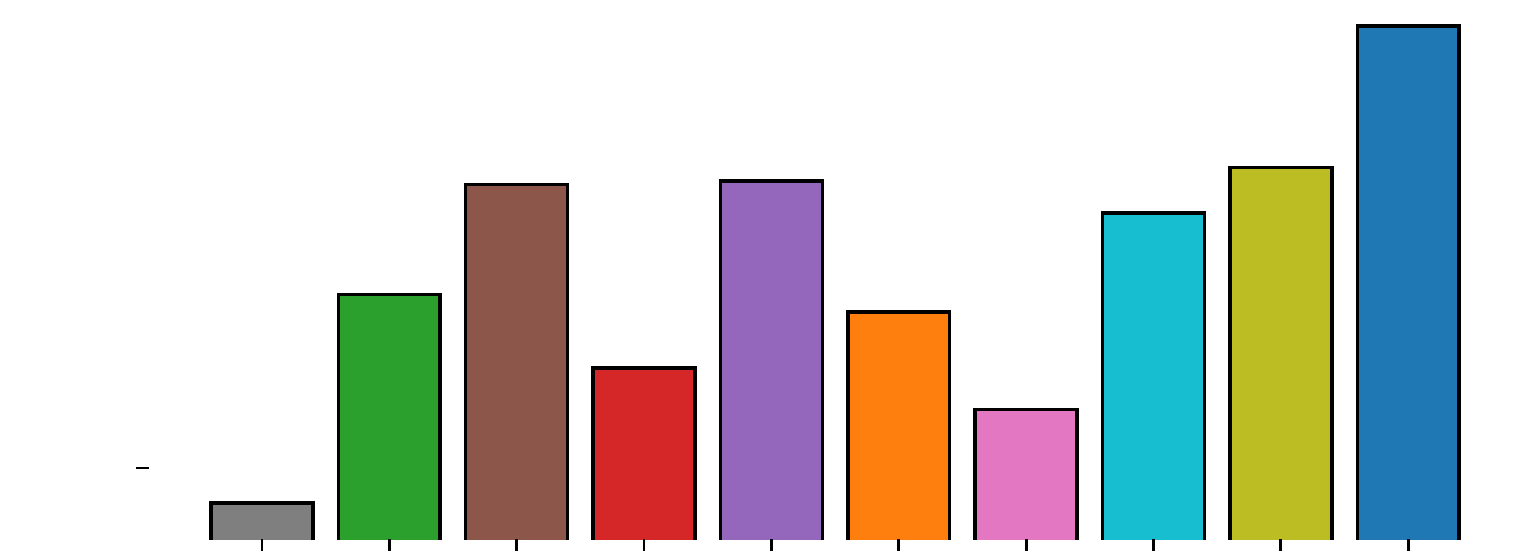}}%
    \put(0.04144226,0.04602924){\makebox(0,0)[lt]{\lineheight{1.25}\smash{\begin{tabular}[t]{l}\scriptsize $10^{9}$\end{tabular}}}}%
    \put(0,0){\includegraphics[width=\unitlength,page=2]{labels.pdf}}%
    \put(0.03118277,0.21309502){\makebox(0,0)[lt]{\lineheight{1.25}\smash{\begin{tabular}[t]{l}\scriptsize $10^{10}$\end{tabular}}}}%
    \put(0,0){\includegraphics[width=\unitlength,page=3]{labels.pdf}}%
    \put(0.01732178,0.18449969){\rotatebox{90}{\makebox(0,0)[t]{\lineheight{1.25}\smash{\begin{tabular}[t]{c}\scriptsize Number of pixels\end{tabular}}}}}%
    \put(0,0){\includegraphics[width=\unitlength,page=4]{labels.pdf}}%
  \end{picture}%
\endgroup%

    \def\svgwidth{\columnwidth}
    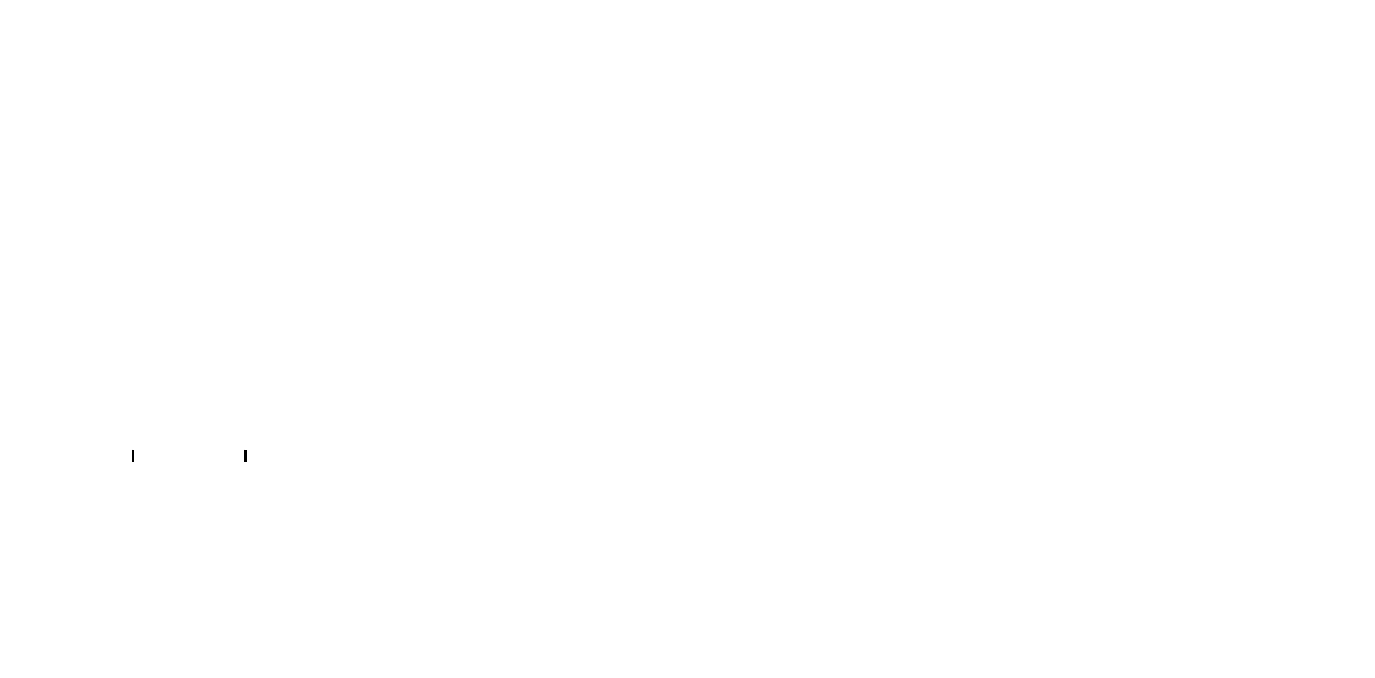
    \caption{Number of pixels per class in the dataset (\textit{top}) and distributions in the images (\textit{bottom}). The boxplot shows the 1st, 2nd (median) and 3rd quartile of the number of pixels in each frame, with the whisker value of 1.5. \textit{background} includes sky and object outside of the garden, while \textit{void} indicates unknown pixels, which should be ignored. }
    \label{fig:label_stat}
\end{figure}

For semantic segmentation, we use the state-of-the-art DeepLabv3+ architecture
with Xception-65 backbone~\cite{Chen2018}. Three aspects of the dataset are
analyzed, namely (1) training size, (2) lighting
conditions, and (3) compatibility with real-world datasets.
The label set is from the 3DRMS challenge~\cite{3drms2017,3drms2018}:
void, grass, ground, pavement, hedge, topiary, flower, obstacle, tree, background.
Background contains the sky and objects outside of the garden, while void indicates
unknown objects to be ignored.
The label statistics are shown in Figure~\ref{fig:label_stat}.
We also follow the network's training setup and report mean
intersection-over-union (mIOU). The results are shown in percentage and higher is better.

\myparagraph{Training and testing size} We first show the benefit of an increasing training set
and the performance on the full and reduced test set. The results are shown in 
Table~\ref{tab:sem_size}.
The performance increases when the training size increases, showing the advantage 
of having large amount of training samples. The evaluation on the reduced test set
is similar to the full set. Thus, unless mentioned otherwise, the test20K 
split will be used for evaluation in later experiments.

\begin{table}[]
    \centering
    \begin{tabular}{@{}ccc@{}}
        \toprule
        \multirow{2}{*}{Sampling} & \multicolumn{2}{c}{test} \\
        \cmidrule(r){2-3} 
        & full & 20K \\
        \midrule
        25\% & 75.71 & 75.89  \\
        50\%  & 79.42 & 79.52 \\
        100\% & 81.96 & 82.09 \\
        \bottomrule
    \end{tabular}
    \caption{Performance with respect to different training size and at 2 test splits.
    The network performance increase when being trained on higher number of images. The
    performance on the reduced test set is on par with the full set.
    }
    \label{tab:sem_size}
\end{table}

\myparagraph{Lighting conditions}
Our dataset contains the same garden models in various lighting conditions,
allowing in-depth analysis of illumination dependency of different methods 
for different tasks. In this section we perform cross-lighting analysis on
semantic segmentation. We conduct lighting-specific training of the networks,
and evaluate the results on each lighting subset of the full test set, as
well as the reduced test set. The results are shown in Table~\ref{tab:lighting_depend}.
All experiments are trained with the same epoch numbers.

For almost all of the categories, training on the specific lighting produces the
best results on that same categories. This is not surprising, as networks always
perform the best on the most similar domains. In general, training with 
cloudy images gives the highest performance, while twilight are the
lowest.  This could be due to relatively bright images and 
less intricate cast shadows in cloudy scenes, in contrast to the mostly dark and
color cast twilight images.

Compared to training with all the full
training set in Table~\ref{tab:sem_size}, the results from training with
lighting-specific images are generally lower and near to the 25\% subset.
This agrees to the training size conclusion as the lighting-specific training
sets account only for around 20\% of the data. Testing on the same lighting
gives a boost in performance, similarly to training with double data size.

\begin{table}[t]
    \centering
    \resizebox{\columnwidth}{!}{
    \setlength{\tabcolsep}{3pt}
    \begin{tabular}{@{}ccccccc@{}}
        \toprule
        \multirow{2}{*}{Training} & \multicolumn{6}{c}{test} \\
        \cmidrule(r){2-7} 
        &  clear &  cloudy &  overcast &  sunset &  twilight &  20K \\
        \midrule
        clear & \bf 76.10 & \emph{76.91} &  76.43 &  72.23 &  75.91 &   72.03 \\
        cloudy &  75.09 &  \bf \emph{77.59} &  77.16 &  72.37 &  76.40 &   \bf 72.30 \\
        overcast &  65.75 &  75.52 &  \bf \emph{78.41} &  70.76 &  74.63 &   70.22 \\
        sunset &  73.21 & 75.76 &  77.17 &  \bf 74.44 &  \emph{77.28} &  71.84 \\
        twilight &  66.19 &  72.86 &  76.21 &  70.55 & \bf \emph{78.16} &  68.83 \\
        \bottomrule
    \end{tabular}
    }
    \caption{Cross-lighting analysis. Each row corresponds to a model trained on the
    specific lighting condition (highest values are in italics), while each column
    corresponds to the results evaluated on the specific subset (highest values are in boldface).
    Lighting-specific training gives better results on the specific lighting, while the results
    in the cross-lighting vary depending on the conditions of the training and test images.}
    \label{tab:lighting_depend}
\end{table}

\myparagraph{Real-world datasets}
Semantic segmentation requires a method to recognize different objects from the 
appearance models learned during training. Therefore, it indicates the closeness of
training data to the testing domain. By analyzing the features learned from EDEN
on real images of unstructured natural scenes, the results indicate the realism
level of our dataset.
To that end, the real-imagery datasets 3DRMS~\cite{3drms2017,3drms2018} (garden scenes, 221 annotated real images for train, 268 for validation, 10 classes),
Freiburg forest~\cite{freiburgforest2016} (forested scenes, 228 annotated real images for train, 15 for validation, 6 classes) are used for evaluation.

The baselines include (1) the network pre-trained on combination of generic datasets,
ImageNet~\cite{imagenet}, COCO~\cite{mscoco}, and augmented PASCAL-VOC
2012~\cite{PascalVOC}, and (2) the network pre-trained on ImageNet and
urban driving scene dataset Cityscapes~\cite{cityscapes2016}.
The encoder part is set to the pre-trained weights provided by the authors~\cite{Chen2018},
while the decoder is finetuned using the train
split of each target set for 50K iterations. The results are shown in
Table~\ref{tab:seg_real}.

The networks using the features learned from EDEN out-perform all alternative
approaches. Both 3DRMS and Freiburg features highly unstructured scenes with
mostly deformable and similar objects found in the nature, drastically different
from the generic images and structured urban scenes. The results show the realism of our dataset to natural scenes and its benefit on training deep networks.
The results on Freiburg test are higher than on 3DRMS due to the relatively simpler and general class labels (\eg. trails, grass, vegetation, and sky) compared to the garden-specific label sets of 3DRMS (\eg hedges, topiaries, roses, tree, \etc).

\begin{table}[t]
    \centering
    \begin{tabular}{@{}ccc@{}}
        \toprule
        \multirow{2}{*}{Pre-training} & \multicolumn{2}{c}{test} \\
        \cmidrule(l){2-3} 
        & 3DRMS & Freiburg \\
        \midrule
        Generic & 24.35 & 41.33 \\
        Cityscapes & 31.11 & 50.08 \\
        EDEN &  \bf 34.55 & \bf 52.45 \\
        \bottomrule
    \end{tabular}
    \caption{Adaptability of features pre-trained on different datasets to
    unstructured natural real-world scenes. The network pre-trained on EDEN
    outperforms all other alternative approaches on both 3DRMS and Freiburg test
    sets.}
    \label{tab:seg_real}
\end{table}

\subsection{Monocular depth prediction}

\begin{figure}[t]
    \centering
    \def\svgwidth{\columnwidth}
    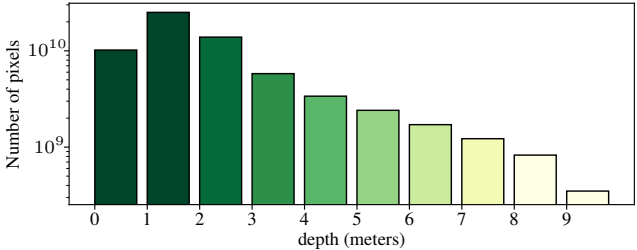
    \caption{Number of pixels per depth range in the dataset. Each range is a left-inclusive half-open interval. }
    \label{fig:depth_stat}
\end{figure}

Monocular depth prediction is an ill-posed problem.
Often the ambiguity is mitigated
by learning from a large-scale depth-annotated
dataset~\cite{Fu2018,Yin2019} or imposing photometric
constraints on image sequences using relative camera poses~\cite{Godard2017,Godard2019}
As camera pose prediction can be formulated using depth constraint,
the depth-pose prediction problems can be combined in a self-supervised learning
pipeline.

Synthetic datasets are favored for being noise-free, which can act as
controlled environments for algorithm analysis. 
In this section, we use EDEN to test different monocular depth prediction networks.
Specifically, we examine the effectiveness of using supervised
signals in learning depth prediction for unstructured natural scenes.
The statistics of the depth in the dataset are shown in Figure~\ref{fig:depth_stat}.

We show the results of training state-of-the-art architectures using different ground 
truth information, namely depth and camera pose.
To that end, the 2 methods, VNL~\cite{Yin2019} and MD2~\cite{Godard2019} are used.
VNL is trained with supervised depth, while MD2 can be trained with ground truth
camera pose or in self-supervised manner. Both are trained using the
schedules and settings provided by the respective authors. The results are shown in 
Table~\ref{tab:depth}. We show the 3 error metrics (rel, log10, rms, seall is better)
after the original work and also include the reported results of the respective methods on the KITTI dataset for comparison.

Generally, supervised method always produce better results than their
self-supervised counterpart as shown by the smaller errors. The difference
are less for the KITTI dataset compared to EDEN. As KITTI contains
mostly rigid objects and
surfaces, it is simpler to obtain predicted camera poses with high accuracy.
On the other hand, camera pose prediction for self-supervised learning on EDEN are
unreliable because of deformable objects and their similarities. The errors are, therefore, 
also higher for supervised methods on EDEN than on KITTI, showing the more challenging
dataset. KITTI has higher RMS numbers due to the larger depth ranges, approximately 80m vs. 15m of EDEN.

\begin{table}[]
    \centering
    \begin{tabular}{@{}ccccccc@{}}
        \toprule
        Method & Supervised & Dataset & rel & log10 & rms \\
        \midrule
        MD2 & None & KITTI & 0.115 & 0.193 & 4.863 \\
        VNL & Depth & KITTI & 0.072 & 0.117 & 3.258 & \\
        \midrule
        MD2 & None & EDEN & 0.438  & 0.556 & 1.403  \\
        MD2 & Pose & EDEN & 0.182 & 0.220 & 0.961 \\
        VNL & Depth & EDEN & 0.181 & 0.083 & 1.061 \\
        \toprule
         & 
    \end{tabular}
    \caption{Performance of different SOTA methods for monocular depth prediction 
    when trained on KITTI and EDEN. The gap is larger between unsupervised and supervised methods on EDEN, showing the necessity of having supervised signals
    for learning unstructured scenes. The errors on EDEN are generally higher than
    on KITTI, showing the more challenging scenes of the (unstructured) dataset.}
    \label{tab:depth}
\end{table}


\section{Conclusion}
\label{sec:conclusion}

The paper presents EDEN, a large-scale multimodal dataset for unstructured
garden scenes, and provides baseline results and analysis on two popular
computer vision tasks, namely the problems of semantic segmentation and
monocular depth prediction.

The experiments show favorable results of using the dataset over generic
and urban-scene datasets for nature-oriented tasks.
The dataset comes with several computer vision modalities and is expected to stimulate applying machine and deep learning to agricultural domains.

\noindent \textbf{Acknowledgements:} This work is performed within the TrimBot2020 project funded by the EU Horizon 2020 program No. 688007.

\section{Add-ons packages}
\label{sec:links}

\begin{itemize}
    \renewcommand{\UrlFont}{\ttfamily\small}
    \item Blender Sapling add-on: \url{https://docs.blender.org/manual/en/latest/addons/add_curve/sapling.html}, Royalty-Free License, last access June, 2020
    \item Arbaro: \url{https://sourceforge.net/projects/arbaro/}, GNU General Public License version 2.0, last access June, 2020
    \item Ivy generator: Thomas Luft, \url{http://graphics.uni-konstanz.de/~luft/ivy_generator}, last access June, 2020
    \item Blender IvyGen add-on: \url{https://docs.blender.org/manual/en/dev/addons/add_curve/ivy_gen.html}, last access June, 2020
    \item Grass essential package: \url{https://blendermarket.com/products/the-grass-essentials}, Royalty-Free License, last access June, 2020.
    \item Grass package: \url{https://www.3d-wolf.com/products/grass.html}, Royalty-Free License, last access June, 2020.
    \item Poliigon: \url{https://www.poliigon.com/}, Royalty-Free license, last access June, 2020.
    \item Pro-Lighting Skies package: \url{https://blendermarket.com/products/pro-lighting-skies},
    \item HDRI Haven: \url{hdrihaven.com}, CC0 license
    \item Essential rock package: \url{https://blendermarket.com/products/the-rock-essentials}, Royalty-Free license
    \item Flower package 1: \url{https://blendermarket.com/products/flowers-pack-1}, Royalty-Free license
    \item Flower package 2: \url{https://blendermarket.com/products/flowers-pack-2}, Royalty-Free license
    \item Garden asset package: \url{https://blendermarket.com/products/garden-asset-pack}, Royalty-Free license
\end{itemize}

{\small
\bibliographystyle{ieee_fullname}
\bibliography{UvA-eden}

\begin{thebibliography}{10}\itemsep=-1pt

\bibitem{Aggarwal1988}
J~K Aggarwal and N Nandhakumar.
\newblock {On the computation of motion from sequences of images-A review}.
\newblock {\em Proceedings of the IEEE}, 76(8):917--935, 1988.

\bibitem{stanford23d2017arxiv}
I Armeni, A Sax, A.$\sim$R. Zamir, and S Savarese.
\newblock {Joint 2D-3D-Semantic Data for Indoor Scene Understanding}.
\newblock {\em ArXiv e-prints}, feb 2017.

\bibitem{middlebury2011}
Simon Baker, Daniel Scharstein, J~P Lewis, Stefan Roth, Michael~J. Black, and
  Richard Szeliski.
\newblock {A Database and Evaluation Methodology for Optical Flow}.
\newblock {\em International Journal of Computer Vision (IJCV)}, 92(1):1--31,
  mar 2011.

\bibitem{Barron1994}
J~L Barron, D~J Fleet, S~S Beauchemin, and T~A Burkitt.
\newblock {Performance of optical flow techniques}.
\newblock {\em International Journal of Computer Vision (IJCV)}, 12(1):43--77,
  feb 1994.

\bibitem{Barth2018}
R Barth, J IJsselmuiden, J Hemming, and E~J van Henten.
\newblock {Data synthesis methods for semantic segmentation in agriculture: A
  Capsicum annuum dataset}.
\newblock {\em Computers and Electronics in Agriculture}, 144:284--296, 2018.

\bibitem{Baslamisli2019arxiv}
Anil~S Baslamisli, Partha Das, Hoang-An Le, Sezer Karaoglu, and Theo Gevers.
\newblock {ShadingNet: Image Intrinsics by Fine-Grained Shading Decomposition}.
\newblock {\em ArXiv e-prints}, 2019.

\bibitem{Baslamisli18eccv}
Anil~S. Baslamisli, Thomas~T. Groenestege, Partha Das, Hoang-An Le, Sezer
  Karaoglu, and Theo Gevers.
\newblock {Joint Learning of Intrinsic Images and Semantic Segmentation}.
\newblock In {\em European Conference on Computer Vision (ECCV)}, jul 2018.

\bibitem{iiw2014}
S Bell, K Bala, and N Snavely.
\newblock {Intrinsic images in the wild}.
\newblock {\em ACM Transactions on Graphics (SIGGRAPH)}, 2014.

\bibitem{camvid2008}
Gabriel~J Brostow, Jamie Shotton, Julien Fauqueur, and Roberto Cipolla.
\newblock {Segmentation and Recognition Using Structure from Motion Point
  Clouds}.
\newblock In {\em Proceedings of the European Conference on Computer Vision
  (ECCV)}, pages 44--57, 2008.

\bibitem{sintel2012}
Daniel~J. Butler, Jonas Wulff, Garrett~B. Stanley, and Michael~J. Black.
\newblock {A naturalistic open source movie for optical flow evaluation}.
\newblock In {A. Fitzgibbon et al. (Eds.)}, editor, {\em Proceedings of the
  European Conference on Computer Vision (ECCV)}, Part IV, LNCS 7577, pages
  611--625. Springer-Verlag, oct 2012.

\bibitem{matterport3d2017}
Angel Chang, Angela Dai, Thomas Funkhouser, Maciej Halber, Matthias Niessner,
  Manolis Savva, Shuran Song, Andy Zeng, and Yinda Zhang.
\newblock {Matterport3D: Learning from RGB-D Data in Indoor Environments}.
\newblock {\em International Conference on 3D Vision (3DV)}, 2017.

\bibitem{shapenet2015}
Angel~X Chang, Thomas Funkhouser, Leonidas Guibas, Pat Hanrahan, Qixing Huang,
  Zimo Li, Silvio Savarese, Manolis Savva, Shuran Song, Hao Su, Jianxiong Xiao,
  Li Yi, and Fisher Yu.
\newblock {ShapeNet: An Information-Rich 3D Model Repository}.
\newblock Technical Report arXiv:1512.03012 [cs.GR], Stanford University ---
  Princeton University --- Toyota Technological Institute at Chicago, 2015.

\bibitem{Chen2018}
Liang-Chieh Chen, Yukun Zhu, George Papandreou, Florian Schroff, and Hartwig
  Adam.
\newblock {Encoder-Decoder with Atrous Separable Convolution for Semantic Image
  Segmentation}.
\newblock In {\em European Conference on Computer Vision (ECCV)}, pages
  833--851. Springer International Publishing, 2018.

\bibitem{cityscapes2016}
Marius Cordts, Mohamed Omran, Sebastian Ramos, Timo Scharw{\"{a}}chter, Markus
  Enzweiler, Rodrigo Benenson, Uwe Franke, Stefan Roth, and Bernt Schiele.
\newblock {The Cityscapes Dataset}.
\newblock In {\em Proceedings of the IEEE Conference on Computer Vision and
  Pattern Recognition (CVPR) Workshops}, volume~3, 2016.

\bibitem{scannet2017}
Angela Dai, Angel~X Chang, Manolis Savva, Maciej Halber, Thomas Funkhouser, and
  Matthias Nie{\ss}ner.
\newblock {ScanNet: Richly-annotated 3D Reconstructions of Indoor Scenes}.
\newblock In {\em Proc. Computer Vision and Pattern Recognition (CVPR), IEEE},
  2017.

\bibitem{imagenet}
J Deng, W Dong, R Socher, L.-J. Li, K Li, and L Fei-Fei.
\newblock {ImageNet: A Large-Scale Hierarchical Image Database}.
\newblock In {\em Proceedings of the IEEE Conference on Computer Vision and
  Pattern Recognition (CVPR)}, 2009.

\bibitem{PascalVOC}
M Everingham, L Van$\sim$Gool, C~K~I Williams, J Winn, and A Zisserman.
\newblock {The Pascal Visual Object Classes (VOC) Challenge}.
\newblock {\em International Journal of Computer Vision (IJCV)},
  88(2):303--338, jun 2010.

\bibitem{Fu2018}
H Fu, M Gong, C Wang, K Batmanghelich, and D Tao.
\newblock {Deep Ordinal Regression Network for Monocular Depth Estimation}.
\newblock In {\em 2018 IEEE/CVF Conference on Computer Vision and Pattern
  Recognition}, pages 2002--2011, 2018.

\bibitem{vkitti2016}
A Gaidon, Q Wang, Y Cabon, and E Vig.
\newblock {Virtual Worlds as Proxy for Multi-Object Tracking Analysis}.
\newblock In {\em Proceedings of the IEEE Conference on Computer Vision and
  Pattern Recognition (CVPR)}, 2016.

\bibitem{kitti2012}
Andreas Geiger, Philip Lenz, and Raquel Urtasun.
\newblock {Are we ready for autonomous driving? The KITTI vision benchmark
  suite}.
\newblock In {\em Proceedings of the IEEE Conference on Computer Vision and
  Pattern Recognition (CVPR)}, pages 3354--3361, 2012.

\bibitem{Godard2017}
Cl{\'{e}}ment Godard, Oisin {Mac Aodha}, and Gabriel~J Brostow.
\newblock {Unsupervised Monocular Depth Estimation with Left-Right
  Consistency}.
\newblock In {\em Proceedings of the IEEE Conference on Computer Vision and
  Pattern Recognition (CVPR)}, 2017.

\bibitem{Godard2019}
Cl{\'{e}}ment Godard, Oisin {Mac Aodha}, Michael Firman, and Gabriel~J Brostow.
\newblock {Digging into Self-Supervised Monocular Depth Prediction}.
\newblock {\em Proceedings of the IEEE International Conference on Computer
  Vision (ICCV)}, oct 2019.

\bibitem{mit2009}
Roger Grosse, Micah~K. Johnson, Edward~H. Adelson, and William~T. Freeman.
\newblock {Ground truth dataset and baseline evaluations for intrinsic image
  algorithms}.
\newblock In {\em Proceedings of the IEEE International Conference on Computer
  Vision (ICCV)}, 2009.

\bibitem{Han2018}
Jian Han, Sezer Karaoglu, Hoang-An Le, and Theo Gevers.
\newblock {Object Features and Face Detection Performance: Analyses with
  3D-Rendered Synthetic Data}.
\newblock In {\em Proceedings of the IEEE International Conference on Pattern
  Recognition (ICPR)}, 2020.

\bibitem{Hewitt2017}
Charlie Hewitt.
\newblock {\em {Procedural Generation of Tree Models for Use in Computer
  Graphics}}.
\newblock PhD thesis, Cambridge Trinity College, 2017.

\bibitem{Horn1981}
Berthold~K.P. Horn and Brian~G. Schunck.
\newblock {Determining optical flow}.
\newblock {\em Artificial Intelligence}, 17(1-3):185--203, aug 1981.

\bibitem{Kaneva2011}
B Kaneva, A Torralba, and W~T Freeman.
\newblock {Evaluation of image features using a photorealistic virtual world}.
\newblock In {\em Proceedings of the IEEE International Conference on Computer
  Vision (ICCV)}, pages 2282--2289, 2011.

\bibitem{Kicanaoglu2018}
B Kicanaoglu, R Tao, and A~W~M Smeulders.
\newblock {Estimating small differences in car-pose from orbits}.
\newblock In {\em Proceedings of the British Machine Vision Conference (BMVC)},
  2018.

\bibitem{saw2017}
Balazs Kovacs, Sean Bell, Noah Snavely, and Kavita Bala.
\newblock {Shading Annotations in the Wild}.
\newblock {\em Proceedings of the IEEE Conference on Computer Vision and
  Pattern Recognition (CVPR)}, 2017.

\bibitem{gtav2018}
P Krahenbuhl.
\newblock {Free Supervision from Video Games}.
\newblock In {\em Proceedings of IEEE Conference on Computer Vision and Pattern
  Recognition (CVPR)}, pages 2955--2964, 2018.

\bibitem{HALe2018}
Hoang-An Le, Anil~S. Baslamisli, Thomas Mensink, and Theo Gevers.
\newblock {Three for one and one for three: Flow, Segmentation, and Surface
  Normals}.
\newblock In {\em Proceedings of the Bristish Machine Vision Conference
  (BMVC)}, jul 2018.

\bibitem{Leibe2007}
B Leibe, N Cornelis, K Cornelis, and L {Van Gool}.
\newblock {Dynamic 3D Scene Analysis from a Moving Vehicle}.
\newblock In {\em 2007 IEEE Conference on Computer Vision and Pattern
  Recognition}, pages 1--8, 2007.

\bibitem{mscoco}
Tsung-Yi Lin, Michael Maire, Serge Belongie, James Hays, Pietro Perona, Deva
  Ramanan, Piotr Dollar, and C~Lawrence Zitnick.
\newblock {Microsoft COCO: Common Objects in Context}.
\newblock In David Fleet, Tomas Pajdla, Bernt Schiele, and Tinne Tuytelaars,
  editors, {\em Proceedings of the European Conference on Computer Vision
  (ECCV)}, pages 740--755, Cham, 2014. Springer International Publishing.

\bibitem{smpl2015}
Matthew Loper, Naureen Mahmood, Javier Romero, Gerard Pons-Moll, and Michael~J
  Black.
\newblock {SMPL: A Skinned Multi-Person Linear Model}.
\newblock {\em ACM Transactions on Graphics (SIGGRAPH)}, 34(6), oct 2015.

\bibitem{flyingthings3d2016}
Nikolaus Mayer, Eddy Ilg, Philip H{\"{a}}usser, Philipp Fischer, Daniel
  Cremers, Alexey Dosovitskiy, and Thomas Brox.
\newblock {A large dataset to train convolutional networks for disparity,
  optical flow, and scene flow estimation}.
\newblock In {\em Proceedings of the IEEE Conference on Computer Vision and
  Pattern Recognition (CVPR)}, pages 4040--4048, 2016.

\bibitem{kitti2015}
Moritz Menze and Andreas Geiger.
\newblock {Object scene flow for autonomous vehicles}.
\newblock In {\em Proceedings of the IEEE Conference on Computer Vision and
  Pattern Recognition (CVPR)}, volume 07-12-June, pages 3061--3070, 2015.

\bibitem{pascalContext}
Roozbeh Mottaghi, Xianjie Chen, Xiaobai Liu, Nam-Gyu Cho, Seong-Whan Lee, Sanja
  Fidler, Raquel Urtasun, Alan Yuille, Raquel Urtasun, and Alan Yuille.
\newblock {The Role of Context for Object Detection and Semantic Segmentation
  in the Wild}.
\newblock In {\em Proceedings of the IEEE Conference on Computer Vision and
  Pattern Recognition (CVPR)}, pages 891--898, 2014.

\bibitem{Mueller2016}
Matthias Mueller, Neil Smith, and Bernard Ghanem.
\newblock {A Benchmark and Simulator for UAV Tracking}.
\newblock In Bastian Leibe, Jiri Matas, Nicu Sebe, and Max Welling, editors,
  {\em European Conference on Computer Vision (ECCV)}, pages 445--461, Cham,
  2016. Springer International Publishing.

\bibitem{mapillary2017}
Gerhard Neuhold, Tobias Ollmann, Samuel {Rota Bul{\`{o}}}, and Peter
  Kontschieder.
\newblock {The Mapillary Vistas Dataset for Semantic Understanding of Street
  Scenes}.
\newblock In {\em Proceedings of the IEEE International Conference on Computer
  Vision (ICCV)}, 2017.

\bibitem{Olszewski2019}
Kyle Olszewski, Sergey Tulyakov, Oliver Woodford, Hao Li, and Linjie Luo.
\newblock {Transformable Bottleneck Networks}.
\newblock In {\em Proceedings of the IEEE International Conference on Computer
  Vision (ICCV)}, nov 2019.

\bibitem{Perlin1985}
Ken Perlin.
\newblock {An Image Synthesizer}.
\newblock {\em SIGGRAPH Computer Graphics}, 19(3):287--296, jul 1985.

\bibitem{viper2017}
Stephan~R Richter, Zeeshan Hayder, and Vladlen Koltun.
\newblock {Playing for Benchmarks}.
\newblock In {\em Proceedings of the IEEE International Conference on Computer
  Vision (ICCV)}, 2017.

\bibitem{Richter2016}
Stephan~R Richter, Vibhav Vineet, Stefan Roth, and Vladlen Koltun.
\newblock {Playing for Data: Ground Truth from Computer Games}.
\newblock In {\em Proceedings of the European Conference on Computer Vision
  (ECCV)}, 2016.

\bibitem{synthia2016}
G Ros, L Sellart, J Materzynska, D Vazquez, A.~M. Lopez, {German Ros;}, {Laura
  Sellart;}, {Joanna Materzynska;}, {David Vazquez;}, and {Antonio M. Lopez;}.
\newblock {The SYNTHIA dataset: A large collection of synthetic images for
  semantic segmentation of urban scenes}.
\newblock In {\em Proceedings of the IEEE Conference on Computer Vision and
  Pattern Recognition (CVPR)}, 2016.

\bibitem{3drms2017}
Torsten Sattler, Radim Tylecek, Thomas Brox, Marc Pollefeys, and Robert~B
  Fisher.
\newblock {3D Reconstruction meets Semantics – Reconstruction Challenge}.
\newblock In {\em Proceedings of the IEEE/CVF International Conference on
  Computer Vision (ICCV) Workshops}, pages 1--7. ICCV Workshops, oct 2017.

\bibitem{nyu2012}
Nathan Silberman, Derek Hoiem, Pushmeet Kohli, and Rob Fergus.
\newblock {Indoor Segmentation and Support Inference from RGBD Images}, 2012.

\bibitem{Taylor2007}
G~R Taylor, A~J Chosak, and P~C Brewer.
\newblock {OVVV: Using Virtual Worlds to Design and Evaluate Surveillance
  Systems}.
\newblock In {\em Proceedings of the IEEE Conference on Computer Vision and
  Pattern Recognition (CVPR)}, pages 1--8, 2007.

\bibitem{3drms2018}
Radim Tylecek, Torsten Sattler, Hoang-An Le, Thomas Brox, Marc Pollefeys,
  Robert~B Fisher, and Theo Gevers.
\newblock {The Second Workshop on 3D Reconstruction Meets Semantics: Challenge
  Results Discussion}.
\newblock In Laura Leal-Taix{\'{e}} and Stefan Roth, editors, {\em Proceedings
  of the European Conference on Computer Vision (ECCV) Workshops}, pages
  631--644, Cham, 2019. Springer International Publishing.

\bibitem{freiburgforest2016}
Abhinav Valada, Gabriel Oliveira, Thomas Brox, and Wolfram Burgard.
\newblock {Deep Multispectral Semantic Scene Understanding of Forested
  Environments using Multimodal Fusion}.
\newblock In {\em International Symposium on Experimental Robotics (ISER)},
  2016.

\bibitem{mapillary2020}
Frederik Warburg, Soren Hauberg, Manuel Lopez-Antequera, Pau Gargallo, Yubin
  Kuang, and Javier Civera.
\newblock {Mapillary Street-Level Sequences: A Dataset for Lifelong Place
  Recognition}.
\newblock In {\em The IEEE/CVF Conference on Computer Vision and Pattern
  Recognition (CVPR)}, jun 2020.

\bibitem{Weber1995}
Jason Weber and Joseph Penn.
\newblock {Creation and rendering of realistic trees}.
\newblock {\em SIGGRAPH '95 - Proceedings of the 22nd Annual Conference on
  Computer Graphics and Interactive Techniques}, pages 119--128, 1995.

\bibitem{Yin2019}
Wei Yin, Yifan Liu, Chunhua Shen, and Youliang Yan.
\newblock {Enforcing geometric constraints of virtual normal for depth
  prediction}.
\newblock In {\em Proceedings of IEEE International Conference on Computer
  Vision (ICCV)}, 2019.

\bibitem{places2014}
Bolei Zhou, Agata Lapedriza, Jianxiong Xiao, Antonio Torralba, and Aude Oliva.
\newblock {Learning Deep Features for Scene Recognition using Places Database}.
\newblock In Z Ghahramani, M Welling, C Cortes, N~D Lawrence, and K~Q
  Weinberger, editors, {\em Advances in Neural Information Processing Systems
  (NIPS)}, pages 487--495. Curran Associates, Inc., 2014.

\bibitem{ade20k}
Bolei Zhou, Hang Zhao, Xavier Puig, Sanja Fidler, Adela Barriuso, and Antonio
  Torralba.
\newblock {Scene Parsing through ADE20K Dataset}.
\newblock In {\em Proceedings of the IEEE Conference on Computer Vision and
  Pattern Recognition (CVPR)}, 2017.

\end{thebibliography}
}

\end{document}